\documentclass[10pt,twocolumn,letterpaper]{article}

\usepackage[pagenumbers]{cvpr} 

\definecolor{cvprblue}{rgb}{0.21,0.49,0.74}
\usepackage[pagebackref,breaklinks,colorlinks,allcolors=cvprblue]{hyperref}
\usepackage{multirow}
\usepackage{tcolorbox}

\def\paperID{10755} 
\def\confName{CVPR}
\def\confYear{2025}

\title{Mitigating Object Hallucinations in Large Vision-Language Models with Assembly of Global and Local Attention\vspace{-4mm}}

\author{Wenbin An$^{1,2}$ \quad
Feng Tian$^{1,2}$\thanks{Corresponding author} \quad
Sicong Leng$^3$ \quad
Jiahao Nie$^{3}$ \quad
Haonan Lin$^{1,2}$ \\
Qianying Wang$^{4}$\footnotemark[1] \quad
Ping Chen$^5$ \quad
Xiaoqin Zhang$^6$ \quad
Shijian Lu$^3$\footnotemark[1]\\
$^1$Xi'an Jiaotong University  \quad
$^2$National Engineering Laboratory for Big Data Analytics \\
$^3$Nanyang Technological University \quad
$^4$Lenovo Research \\
$^5$University of Massachusetts Boston  \quad
$^6$Zhejiang University of Technology \\
{\tt\small \{wenbinan,linhaonan\}@stu.xjtu.edu.cn,
fengtian@mail.xjtu.edu.cn, wangqya@lenovo.com} \\
{\tt\small 
\{jiahao007,sicong001\}@e.ntu.edu.sg, Ping.Chen@umb.edu,  shijian.lu@ntu.edu.sg}
}

\begin{document}
\maketitle

\begin{abstract}
Despite great success across various multimodal tasks, Large Vision-Language Models (LVLMs) often encounter object hallucinations with generated textual responses being inconsistent with the actual objects in images. We examine different LVLMs and pinpoint that one root cause of object hallucinations lies with \textit{deficient attention} on discriminative image features. Specifically, LVLMs often predominantly attend to prompt-irrelevant global features instead of prompt-relevant local features, undermining their visual grounding capacity and leading to object hallucinations. We propose \textit{Assembly of Global and Local Attention (AGLA)}, a training-free and plug-and-play approach that mitigates hallucinations by assembling global features for response generation and local features for visual discrimination simultaneously. 
Specifically, we introduce an image-prompt matching scheme that captures prompt-relevant local features from images, leading to an augmented view of the input image where prompt-relevant content is highlighted while irrelevant distractions are suppressed. 
Hallucinations can thus be mitigated with a calibrated logit distribution that is from generative global features of the original image and discriminative local features of the augmented image.
Extensive experiments show the superiority of AGLA in LVLM hallucination mitigation, demonstrating its wide applicability across both discriminative and generative tasks.
Our code is available at \href{https://github.com/Lackel/AGLA}{https://github.com/Lackel/AGLA}.
\end{abstract}
\vspace{-5mm}    
\section{Introduction}
By extending Large Language Models~\cite{brown2020language,touvron2023llama,bai2023qwenllm,vicuna2023,alpaca} into the visual domain, Large Vision-Language Models (LVLMs)~\cite{liu2023visual,zhu2023minigpt4,ye2023mplugowl,li2023otter,dai2023instructblip,gong2023multimodalgpt,bai2023qwen} have demonstrated impressive performance across various tasks such as image captioning~\cite{li2023blip,wang2023caption} and visual question answering~\cite{lee2024visual,wang2024weakly,an2024knowledge}. 
Despite their remarkable success, LVLMs are still facing several challenges that limit their reliability and applicability in various vision-language tasks. In particular, one prominent challenge lies with object hallucinations~\cite{liu2023mitigating,lovenia2023negative,vcd,liu2024survey, deng2024seeing,zhu2024ibd} where the LVLM generated textual responses are inconsistent with the actual objects in the input images~\cite{gunjal2023detecting,li2023evaluating}.

\begin{figure}[t]
    \centering
    \vspace{0mm}
    \includegraphics[width=\linewidth]{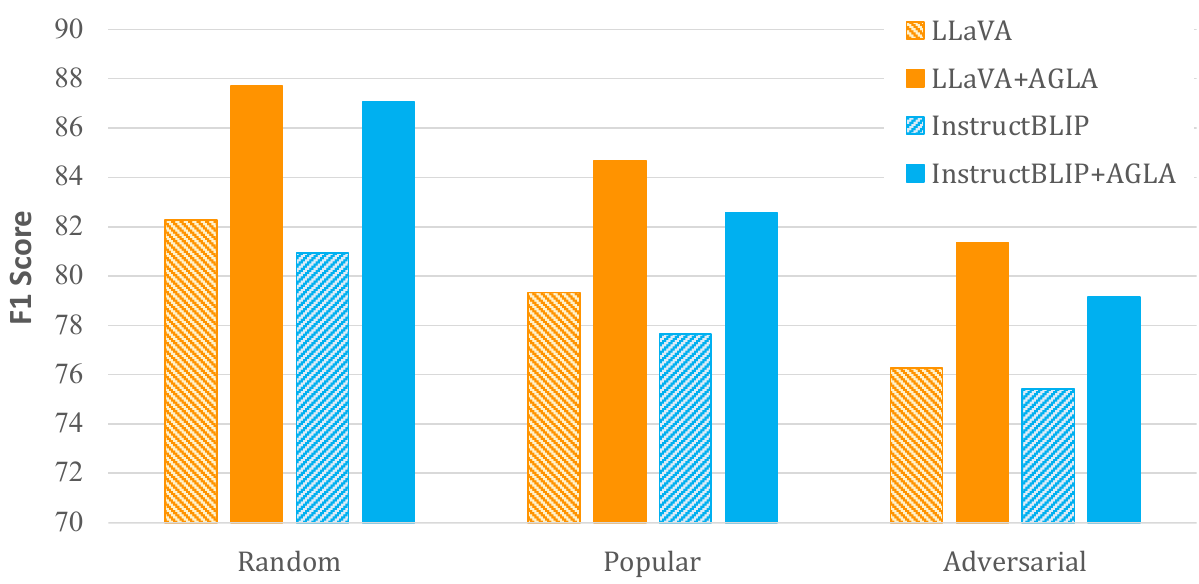}
    \vspace{-7mm}
    \captionof{figure}{Efficacy of \textit{AGLA} under different settings of the POPE dataset~\cite{li2023evaluating}. A lower F1 score means a higher hallucination rate.}
    \label{fig:intro}
    \vspace{-4mm}
\end{figure}

\begin{figure*}[t]
\vspace{-6mm}
\centering
    \includegraphics[width=0.93\textwidth]{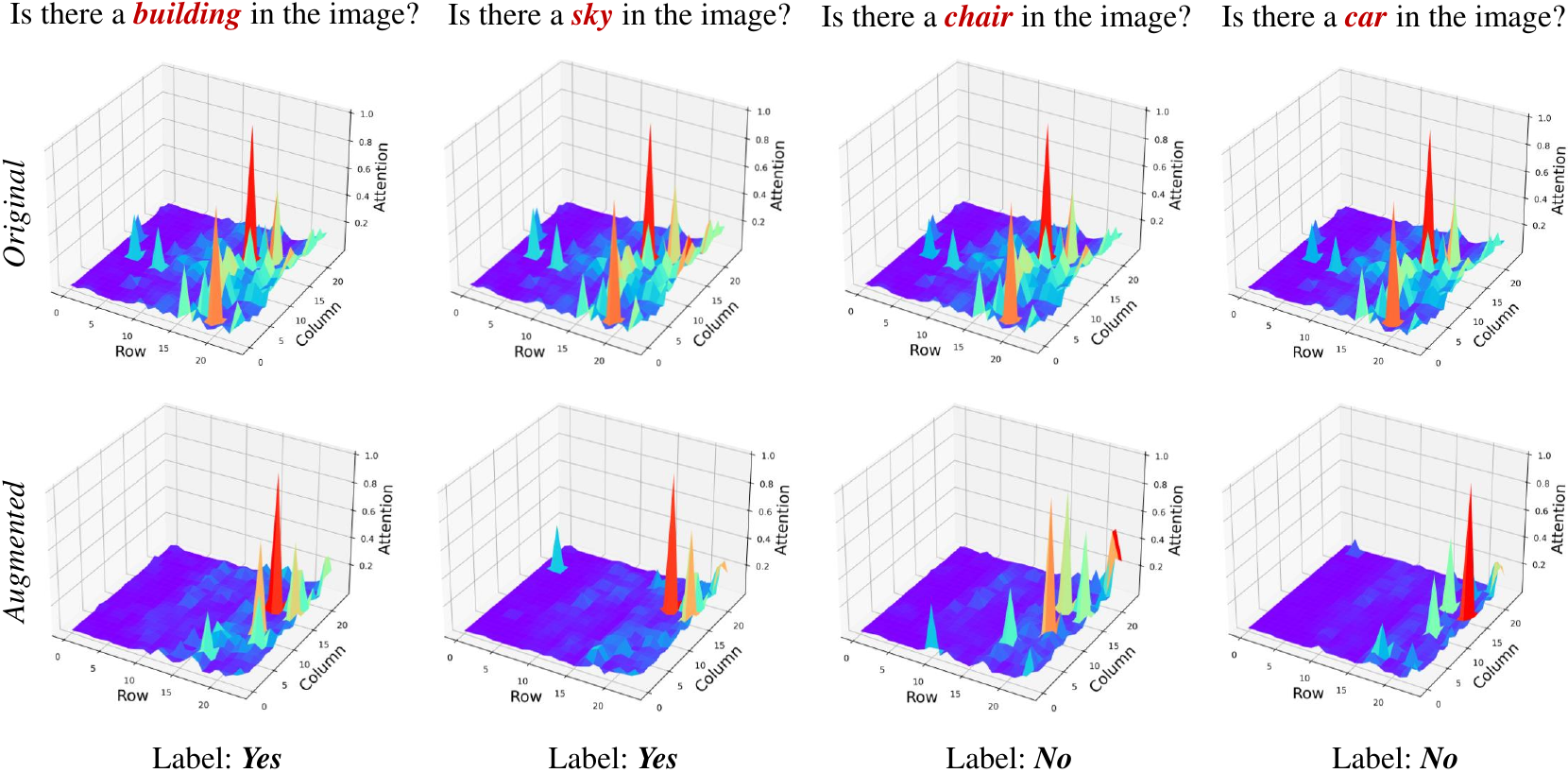}
    \vspace{-2mm}
    \captionof{figure}{The weights of LVLM self-attention with respect to image patch features when LVLM responds to different object queries (\textit{Yes} or \textit{No}). The \textbf{first row} shows the LVLM self-attention toward the original image, where the attention is dominated by certain global features and demonstrates similar patterns consistently regardless of the queried objects. The \textbf{second row} shows the LVLM self-attention toward augmented views of the image, where the attention is more prompt-relevant and captures query-relevant local features.}
    \label{fig:attention}
\vspace{-3mm}
\end{figure*}

Several studies have explored the underlying cause of object hallucinations in LVLMs, leading to different findings such as statistical pre-training bias~\cite{agarwal2020towards, agrawal2016analyzing,goyal2017making}, over-reliance on parametric knowledge~\cite{vcd,lee2023volcano,zhibo2023overcoming,han2022visual,wu2022overcoming}, biased feature learning~\cite{zhu2024ibd,huang2023opera,yue2024less}, etc.
Despite these efforts, we conjecture that object hallucinations stem from certain \textit{attention deficiency} toward images, where LVLMs predominantly attend to global image features while failing to capture prompt-relevant local features.
To verify this hypothesis, we examine the weights of LVLM self-attention toward image features while LVLMs respond to text queries about different objects. As the first row of Fig.~\ref{fig:attention} shows, the weights of LVLM self-attention are dominated by certain global features and present similar patterns regardless of the queried objects being present or absent in the image. This study demonstrates how the \textit{attention deficiency} in LVLMs causes the ignorance of prompt-relevant object features in images~\cite{kabbai2019image} and how it impairs LVLMs’ visual grounding ability and leads to object hallucinations. Such a tendency toward global image features is well aligned with the recent studies~\cite{park2023self,agarwal2020towards,vcd}. To verify this hypothesis, we further conduct experiments on different settings of the POPE dataset~\cite{li2023evaluating}. As Fig.~\ref{fig:intro} shows, LVLMs are more likely to hallucinate objects that co-appear frequently in images (\ie, the adversarial setting), showing that LVLMs tend to be disturbed by prompt-irrelevant features through association (\eg, hallucinating ``cars'' disturbed by ``road'' in images).
It is thus crucial to incorporate prompt-relevant local attention to highlight prompt-relevant features and suppress prompt-irrelevant distractions for better hallucination mitigation.

To this end, we design the \textit{Assembly of Global and Local Attention} (\textit{AGLA}), a training-free and plug-and-play approach that introduces prompt-relevant local attention to generate an augmented view of the image for hallucination mitigation. Inspired by the idea of network interpretability~\cite{selvaraju2017grad,tiong2022plug,sun2024review}, we design an image-prompt matching scheme that computes the prompt-relevant attention and further generates an augmented view of the original image. As illustrated in the second row of Fig.~\ref{fig:attention}, the LVLM self-attention over the augmented image responds well to the object queries, which forms a great complement to the LVLM self-attention over the original input image. 
We thus integrate the generative global features from the original image and the discriminative local features from the augmented image, leading to a calibrated logit distribution that enhances the perception capabilities of LVLMs and mitigates object hallucinations. 
Extensive experiments over multiple generative and discriminative benchmarks show that AGLA achieves superior performance on hallucination mitigation as well as other perception tasks, demonstrating its superior reliability under various challenging scenarios.

The major contributions of this work can be summarized in three aspects:
\begin{itemize}
    \item We examine object hallucinations in LVLMs and identify \textit{attention deficiency} as one major cause. This finding offers a new perspective and opens a new door for understanding and mitigating hallucinations. 
    \item We design AGLA, a training-free and plug-and-play decoding approach that ensembles prompt-irrelevant global attention and prompt-relevant local attention to capture both generative and discriminative image features, greatly enhancing the perception ability of LVLMs.
    \item Extensive experiments over multiple generative and discriminative benchmarks show that AGLA achieves superior performance in hallucination mitigation.
\end{itemize}
\section{Related Work}
\subsection{Large Vision-Language Models}
Benefited from the rapid development of Large Language Models (LLMs)~\cite{gilardi2023chatgpt,touvron2023llama,tay2022ul2,raffel2020exploring,brown2020language,chowdhery2022palm,alpaca,vicuna2023,bai2023qwenllm}, Large Vision-Language Models (LVLMs)~\cite{liu2023visual,dai2023instructblip,zhu2023minigpt4,li2023otter,ye2023mplugowl,bai2023qwen} have made great strides by incorporating visual encoders and feature projectors~\cite{li2019visualbert,sun2019videobert,wang2022git,li2022blip}, marking a significant enhancement in their performance and adaptability on various multimodal tasks~\cite{li2023blip,wang2023caption,lee2024visual,wang2024weakly,lin2024dreamsalon}.
Most of existing LVLMs share the same two training phases, i.e., pre-training for feature alignment and instruction-based LLM fine-tuning~\cite{li2019visualbert,li2022blip,dai2023instructblip}. Recently, several studies have attempted to enhance the alignment~\cite{shu2025large} between model responses and human values through Reinforcement Learning from Human Feedback (RLHF)~\cite{sun2023aligning} and preference fine-tuning~\cite{zhou2024aligning}. Nevertheless, similar to LLMs, LVLMs are prone to generating hallucinations~\cite{chair,wang2023evaluation,nie2024mmrel,leng2024curse,zhang2024seeing}, which significantly impacts their reliability and stability for real-world applications.

\subsection{Object Hallucination}
Hallucination~\cite{ji2023survey,zhang2023siren,shi2023replug} in the NLP community is defined as the generation of incorrect or nonsensical content. 
Recently, this issue has also attracted increasing attention in the multimodal domain. Among different types of multimodal hallucinations~\cite{liu2024survey}, ``object hallucination'' is most precisely defined and widely studied, referring to LVLMs generating responses that include objects not present in images~\cite{chair,biten2022let,li2023evaluating,wang2023evaluation,li2023evaluating,lovenia2023negative}. Several studies have demonstrated that object hallucination may come from the statistical bias of pre-training~\cite{agarwal2020towards,agrawal2016analyzing,goyal2017making}, over-reliance on parametric knowledge~\cite{vcd,lee2023volcano,zhibo2023overcoming,han2022visual,wu2022overcoming} and biased feature learning~\cite{zhu2024ibd,huang2023opera,yue2024less}.  Despite these efforts, more underlying reasons for object hallucinations are still under-explored. 

To mitigate hallucinations, previous work usually introduces instruction tuning~\cite{gunjal2023detecting,li2023m,liu2023aligning,liu2023mitigating,chen2023mitigating,cca}, trains a post-hoc reviser~\cite{zhou2023analyzing,yin2023woodpecker,wu2024logical} or designs decoding strategies~\cite{huang2023opera,vcd,zhu2024ibd,chuang2023dola}. For example, Woodpecker~\cite{yin2023woodpecker} utilizes ChatGPT~\cite{brown2020language} to revise the generated responses.
DOLA~\cite{chuang2023dola} contrasts the differences in logits obtained from the later and earlier layers of LLMs. OPERA~\cite{huang2023opera} proposes a penalty-based decoding method equipped with a retrospection-reallocation strategy to penalize candidates that may lead to hallucinations. VCD~\cite{vcd} extends contrastive decoding in NLP~\cite{li2022contrastive,o2023contrastive} into the multimodal domain by contrasting outputs from original and distorted images. Nevertheless, these methods struggle to capture fine-grained local image features, resulting in poor grounding capabilities and object hallucinations in LVLMs.
\section{Method}

To direct LVLMs' focus to image regions relevant to prompts and mask out distractions, we propose an \textit{Image-Prompt Matching (IPM)} technique to generate an augmented view of the input image (Sec.~\ref{sec41}). Building on this, we introduce the \textit{Assembly of Global and Local Attention} (\textit{AGLA}), which integrates generative global features from original images with discriminative local features from augmented images through logit fusion, thereby deriving a calibrated distribution for more accurate decoding (Sec.~\ref{sec42}).


\subsection{Image-Prompt Matching}
\label{sec41}

\begin{figure}[t]
\centering
    \vspace{-2mm}
    \includegraphics[width=0.98\linewidth]{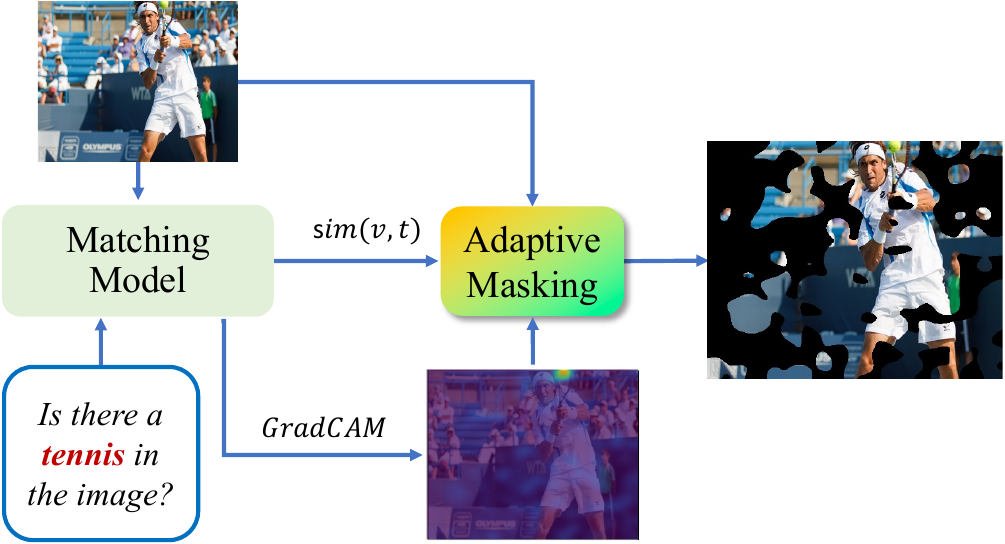}
    \vspace{-2.5mm}
    \caption{An illustration of the proposed Image-Prompt Matching. $sim(\textit{v}, \textit{t})$ is an output score of the matching model, which measures the similarity between image \textit{v} and prompt \textit{t}.}
    \vspace{-3.5mm}
    \label{fig:model2}
\end{figure}

Large Vision-Language Models (LVLMs) often fail to focus on image regions relevant to input prompts due to \textit{attention deficiency}. To address this issue, we introduce an \textit{Image-Prompt Matching} (\textit{IPM}) module (illustrated in Fig.~\ref{fig:model2}) to identify and mask out irrelevant image regions, ensuring that LVLMs focus on prompt-relevant content. This module first calculates an overall similarity score, $sim(\textit{v}, \textit{t})$, between the image \textit{v} and the textual prompt \textit{t} based on a matching model. Leveraging advancements in interpretability techniques, we apply GradCAM~\cite{selvaraju2017grad,tiong2022plug} to the cross-attention layer of the matching model, yielding a correlation score for each image patch relative to the input prompts. Specifically, let $X \in \mathbb{R}^{M \times D_t}$ represent the prompt token features and $Y \in \mathbb{R}^{K \times D_v}$ represent the image patch features ($M$ and $K$ denote the number of prompt tokens and image patches, respectively). The cross-attention matrix $C \in \mathbb{R}^{M \times K}$ can be computed as follows:
\begin{equation}
    C = \operatorname{softmax}\left(\frac{X W_T W_V^\top Y^\top}{\sqrt{D_t}}\right)
\end{equation}
where $W_T \in \mathbb{R}^{D_t \times D_t}$ and $W_V \in \mathbb{R}^{D_v \times D_t}$ denote parameters of the cross-attention heads. $D_t$ and $D_v$ denote the dimensions of prompt token and image patch features, respectively. The cross-attention matrix $C$ quantifies the attention each prompt token allocates to each image patch, with $C_{ij}$ representing the attention between the $i$-th prompt token and the $j$-th image patch. The correlation score for the $j$-th image patch with respect to the entire textual prompt can thus be computed as follows:

\begin{equation}
\text{cor}(j) = \frac{1}{H} \sum^{M}_{i=1} \sum^{H}_{h=1} \max\left(0,\frac{\partial  \text{ sim}(v, t)}{\partial C_{ij}^{(h)}}\right) C_{ij}^{(h)}
\end{equation}
where $H$ denotes the number of cross-attention heads, and $C^{(h)}$ is the cross-attention matrix for the $h$-th head. The partial derivative term measures the sensitivity of the similarity score to the cross-attention score, indicating the importance of each attention score. This process allows computing the correlation score for each image patch and identifying those most relevant to the input prompts.

To ensure LVLMs focus on prompt-relevant image content while blocking out distraction objects, we introduce an adaptive masking strategy based on the computed correlation scores. This strategy masks out image regions with low correlation scores while retaining those with high scores.
We determine the ratio of masking adaptively based on the overall similarity score $sim(\textit{v}, \textit{t})$, instead of using a manually-defined ratio for all images and prompts. Specifically, we simply adopt $sim(\textit{v}, \textit{t})/2$ as the masking ratio, allowing the masking process to adapt to varying levels of image-prompt matching. For image-prompt pairs with higher similarity scores, indicating a higher relevance between the image and prompt, a larger portion of the image will be masked out for effective distraction suppression and hallucination mitigation.

\begin{figure}[t]
\centering
\vspace{-5mm}
    \includegraphics[width=\linewidth]{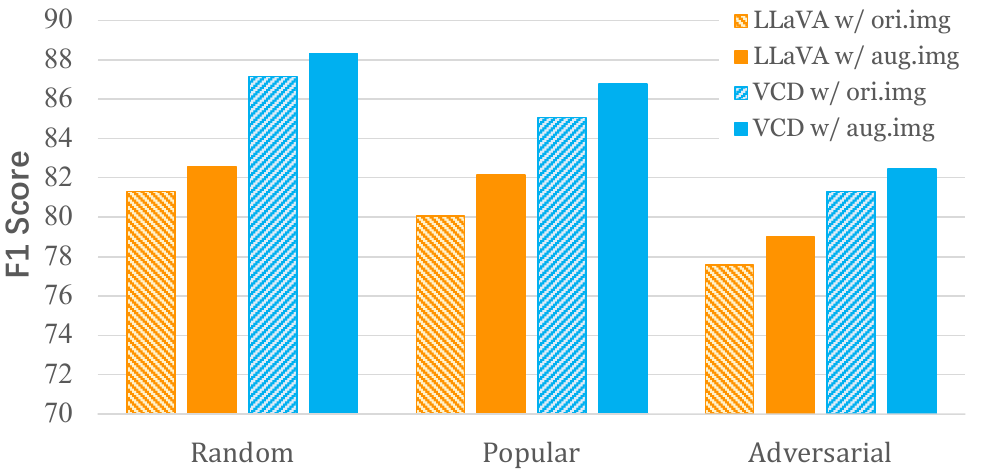}
    \vspace{-7mm}
    \captionof{figure}{Performances with original or augmented input images.}
    \label{fig:aug}
\vspace{-1.5mm}
\end{figure}

\noindent \textbf{Validation.} 
With the help of the matching model, LVLMs can focus on image regions corresponding to the prompt and capture prompt-dependent local features, thereby alleviating the \textit{attention deficiency} issue.
To validate the effectiveness of the proposed IPM module, we perform experiments on the POPE~\cite{li2023evaluating} dataset. As shown in Fig.~\ref{fig:aug}, LLaVA~\cite{liu2023visual} and VCD~\cite{vcd} both perform better by simply replacing original images with the augmented images from IPM, demonstrating the effectiveness of IPM in distraction suppression and hallucination mitigation.

\subsection{Assembly of Global and Local Attention}
\label{sec42}

\begin{figure}[t]
    \centering
    \vspace{-5mm}
    \includegraphics[width=\linewidth]{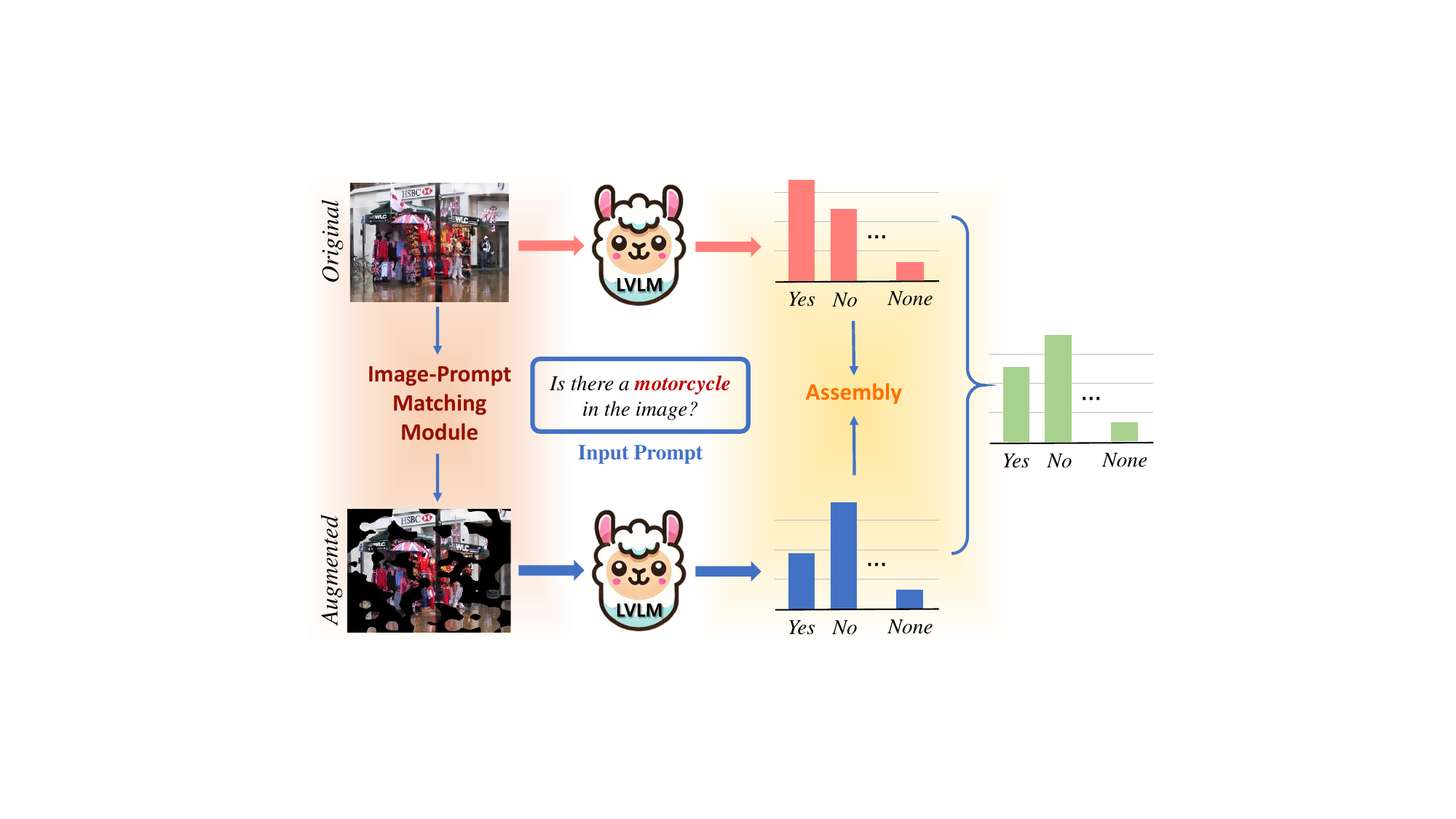}
    \vspace{-4mm}
    \captionof{figure}{Decoding with \textit{Assembly of Global and Local Attention}.}
    \label{fig:abc}
    \vspace{-5mm}
\end{figure}

While IPM effectively blocks out distractions and captures crucial local features for visual discrimination, it inevitably introduces information loss because it blocks out some global features that are essential for generative tasks. To address this, we design AGLA that integrates local and global attention to capture both discriminative and generative features, thereby learning more comprehensive image representations. Specifically, as Fig.~\ref{fig:abc} shows, at each decoding step $i$, we assemble the logits derived from both the original and augmented images to obtain a calibrated decoding distribution. This strategy can be formulated as follows:
\begin{equation}
\label{fuse}
\begin{gathered}
p_{AGLA}(y_i|v,v^{aug},t,y_{<i}) \sim \operatorname{softmax}[\operatorname{logit}_\theta(y_i \mid v, t, y_{<i}) \\
+ \alpha \operatorname{logit}_\theta\left(y_i \mid v^{aug}, t, y_{<i}\right)]
\end{gathered}
\end{equation}
where $y_i$ denotes the token at decoding step $i$, and $y_{<i}$ represents the sequence of generated tokens before step $i$. The variables $v$, $v^{aug}$, and $t$ denote the original image, the augmented image, and the input prompt, respectively. The parameter $\theta$ denotes the model parameters of the LVLMs, and $\alpha$ is a weighting coefficient that balances the contributions of local and global logits. Different from VCD~\cite{vcd} which models a noise distribution and subtracts the distribution from the original one, our model generates a useful distribution that focuses on local regions of interest and adds this distribution to the original one as a supplement to mitigate the \textit{attention deficiency} issue. The two approaches are thus orthogonal and complementary to each other.

\noindent \textbf{Adaptive Plausibility Constraints.} 
Previous works~\cite{li2022contrastive,vcd,zhu2024ibd} suggest that calibrating the entire output distribution, as in Eq.~\ref{fuse}, may penalize valid outputs from the original distribution and promote implausible outputs from the augmented distribution. To mitigate this issue, we adopt adaptive plausibility constraints~\cite{li2022contrastive} to selectively consider tokens with high original probabilities and truncate other tokens as follows:
\begin{equation}
\label{constraint}
\begin{gathered}
\begin{aligned}
&\mathcal{V}_{\text {token}}\left(y_{<i}\right)= 
 \{y_i \in \mathcal{V}: \\
 &p_{\theta}\left(y_i \mid v,t,y_{<i}\right) \geq \beta \max _w p_{\theta}\left(w \mid v,t,y_{<i}\right)\}
\end{aligned}
\\
\begin{aligned}
p_{AGLA}\left(y_i \mid v, v^{aug}, t, y_{<i}\right) = 0, \text{ if } y_i \notin \mathcal{V}_{\text {token}}\left(y_{<i}\right)
\end{aligned}
\end{gathered}
\end{equation}
where $\mathcal{V}_{\text{token}}$ is the set of selected tokens and $\mathcal{V}$ is the output vocabulary. $\beta$ is a hyper-parameter that controls the strength of truncation. A larger $\beta$ means only high-probability tokens will be retained. Effects of hyper-parameters (\ie, $\alpha$ and $\beta$) are discussed in Appendix \ref{app_hyper}.

\begin{table*}[t]
\centering
\vspace{-6mm}
\caption{Experimental results on the three POPE subsets with LLaVA-1.5~(7B) and InstructBLIP~(7B).}
\vspace{-2mm}
\renewcommand\arraystretch{1.02}
\scalebox{1}{%
\setlength{\tabcolsep}{12.5pt}
\begin{tabular}{cclccc|c}
\hline\hline
\textbf{Model}                         & \textbf{Setting}     & \textbf{Decoding} & Accuracy $\uparrow$ & Precision & Recall & F1 Score $\uparrow$  \\ \hline\hline
\multirow{15}{*}{LLaVA-1.5~\cite{liu2023visual}}      
& \multirow{5}{*}{\textit{Random}}     
& Regular~\cite{vcd}           &83.49 &88.84 &76.76 &82.28  \\& 
& DOLA~\cite{chuang2023dola}              &84.78 &87.59 &81.27 &84.19  \\&
& OPERA~\cite{huang2023opera}             &87.53 &94.52 &79.80 &86.45  \\&
& VCD~\cite{vcd}               &86.84 &87.15 &86.68 &86.83  \\& 
& \textbf{AGLA}               &\textbf{88.54} &94.41 &82.08 &\textbf{87.71}  \\              
\cline{2-7}

& \multirow{5}{*}{\textit{Popular}}     
& Regular~\cite{vcd}           &79.98 &82.47 &76.76 &79.34  \\& 
& DOLA~\cite{chuang2023dola}              &79.75 &84.11 &76.22 &80.61  \\&
& OPERA~\cite{huang2023opera}             &84.21 &88.00 &79.80 &83.50  \\&
& VCD~\cite{vcd}               &82.65 &87.15 &80.59 &83.37  \\&          
& \textbf{AGLA}               &\textbf{85.14} &87.88 &82.08 &\textbf{84.68}  \\
\cline{2-7}

& \multirow{5}{*}{\textit{Adversarial}}     
& Regular~\cite{vcd}           &76.03 &76.11 &76.80 &76.26  \\&
& DOLA~\cite{chuang2023dola}              &76.32 &77.27 &75.47 &76.16  \\&
& OPERA~\cite{huang2023opera}             &80.88 &82.16 &79.76 &80.69  \\&
& VCD~\cite{vcd}               &77.31 &73.43 &86.47 &79.28  \\&          
& \textbf{AGLA}               &\textbf{81.13} &81.20 &82.10 &\textbf{81.36}  \\
\hline\hline

\multirow{15}{*}{InstructBLIP~\cite{dai2023instructblip}}

& \multirow{5}{*}{\textit{Random}} 
& Regular~\cite{vcd}           &80.42 &78.93 &83.21 &80.94  \\&
& DOLA~\cite{chuang2023dola}              &83.00 &83.06 &83.13 &83.00  \\&
& OPERA~\cite{huang2023opera}             &85.07 &88.39 &80.73 &84.39  \\&
& VCD~\cite{vcd}               &84.11 &84.20 &85.36 &84.13  \\&          
& \textbf{AGLA}               &\textbf{87.30} &88.83 &85.68 &\textbf{87.07}  \\
\cline{2-7}


& \multirow{5}{*}{\textit{Popular}} 
& Regular~\cite{vcd}           &76.09 &73.22 &82.94 &77.65  \\&  
& DOLA~\cite{chuang2023dola}              &78.99 &77.12 &83.13 &79.85  \\&
& OPERA~\cite{huang2023opera}             &78.33 &73.85 &87.73 &80.20  \\&
& VCD~\cite{vcd}               &79.94 &77.84 &83.33 &80.80  \\&
& \textbf{AGLA}               &\textbf{81.86} &80.17 &85.68 &\textbf{82.58}  \\
\cline{2-7}

& \multirow{5}{*}{\textit{Adversarial}} 
& Regular~\cite{vcd}           &72.37 &68.78 &83.06 &75.42  \\&
& DOLA~\cite{chuang2023dola}              &74.67 &71.53 &83.11 &76.68  \\&
& OPERA~\cite{huang2023opera}             &75.50 &70.49 &87.73 &78.17 \\&
& VCD~\cite{vcd}               &76.32 &73.24 &84.08 &78.08  \\&          
& \textbf{AGLA}               &\textbf{77.29} &74.09 &85.67 &\textbf{79.16}  \\
\hline\hline
\end{tabular}
}
\label{tab:average}
\vspace{-2mm}
\end{table*}

\noindent \textbf{Understanding from the Temperature Perspective.} The proposed integration strategy can be interpreted as applying adaptive sampling temperatures~\cite{holtzman2019curious} during decoding. For similar predictions from the original and augmented images, the integration effectively sets a lower temperature for the original logits, skewing the distribution towards high-probability tokens and avoiding the selection of low-probability tokens. Conversely, for dissimilar predictions, the integration works like setting a higher temperature, making the distribution closer to uniform, and thus increasing the likelihood of sampling correct tokens by giving more balanced consideration to various tokens.

\section{Experiments}
\subsection{Experimental Settings}
\label{setting}

\noindent \textbf{Datasets.} We evaluate our model on both discriminative and generative datasets. More details about the datasets and evaluation metrics are provided in Appendix~\ref{app_metric}.

\noindent \textbf{POPE}: The Polling-based Object Probing Evaluation~\cite{li2023evaluating} contains 27,000 Yes/No questions about \textbf{object existence} in three datasets: MSCOCO~\cite{lin2014microsoft}, A-OKVQA~\cite{schwenk2022okvqa}, and GQA~\cite{hudson2019gqa}. Each dataset includes three negative sample settings: random, popular, and adversarial. The evaluation metrics include Accuracy, Precision, Recall, and F1 score.

\noindent \textbf{ROPE}: The Recognition-based Object Probing Evaluation~\cite{chen2024multiobject} assesses \textbf{multi-object hallucination}, which can evaluate models in real-world complex queries that often involve multiple objects. The evaluation metrics include Accuracy, Precision, Recall, and F1 score.

\noindent \textbf{MME}: This comprehensive benchmark~\cite{fu2023mme} assesses the overall ability of LVLMs such as \textbf{attributes} and \textbf{relationships} of objects, with performance evaluated via the total score of Accuracy and Accuracy+.

\noindent \textbf{CHAIR}: Caption Hallucination Assessment with Image Relevance (CHAIR)~\cite{chair} quantifies object hallucinations in \textbf{image captions} by comparing generated objects to ground-truth ones. Following previous works~\cite{huang2023opera,yue2024less}, we randomly select 500 images from MSCOCO~\cite{lin2014microsoft} and use official CHAIR$_{I}$, CHAIR$_{S}$, and Recall as evaluation metrics.

\noindent \textbf{LLaVA-Bench-Wild}: This generative dataset~\cite{liu2023visual} contains 24 images with 60 questions to assess the capability of LVLMs in tackling challenging tasks and their adaptability to \textbf{new domains}. Following previous work~\cite{vcd,huang2023opera}, we use GPT-4 to evaluate the accuracy and detailedness of generated captions. More details are shown in Appendix~\ref{app_gpt4}.

\begin{table*}[tp]
\vspace{-5mm}
\caption{Experimental results on the ROPE dataset with LLaVA-1.5~(7B) and MiniCPM-V~(2.4B).}
\vspace{-2mm}
\centering
\renewcommand\arraystretch{1.02}
\scalebox{1}{
\setlength{\tabcolsep}{11.3pt}
\begin{tabular}{cllccc|c}
\hline\hline
\textbf{Model}          & \textbf{Setting}                  & \textbf{Decoding} & Accuracy $\uparrow$  & Precision & Recall & F1 Score $\uparrow$   \\ \hline\hline
\multirow{10}{*}{LLaVA-1.5~\cite{liu2023visual}}  & \multirow{2}{*}{\textit{Adversarial-A}}           & Regular~\cite{liu2023visual}           &10.69 &38.51 &14.82 &21.40                                         \\&                               & AGLA               &\textbf{43.77} &49.47 &43.77 &\textbf{46.45}  \\
                          \cline{2-7} 
                          & \multirow{2}{*}{\textit{Adversarial-B}}     & Regular~\cite{liu2023visual}           &9.88 &37.09 &13.92 &20.24  \\
                                    &                               & AGLA               &\textbf{44.35} &50.61 &44.61 &\textbf{47.42}   \\

                        \cline{2-7}
                          & \multirow{2}{*}{\textit{Heterogeneous}} &  Regular~\cite{liu2023visual}           &2.80 &9.46 &4.00 &5.62  \\
                                   &                               & AGLA               &\textbf{7.72} &13.00 &7.72 &\textbf{9.69}  \\
                        \cline{2-7}
                          & \multirow{2}{*}{\textit{Homogenous}} &  Regular~\cite{liu2023visual}           &18.45 &59.59 &25.15 &35.37  \\
                                   &                               & AGLA               &\textbf{60.45} &70.81 &60.92 &\textbf{65.49}  \\
                        \cline{2-7}
                          & \multirow{2}{*}{\textit{Mixed}} &  Regular~\cite{liu2023visual}           &7.02 &23.54 &9.82 &13.85  \\
                                   &                               & AGLA               &\textbf{27.15} &31.81 &27.33 &\textbf{29.40}  \\
                                   \hline\hline
\multirow{10}{*}{MiniCPM-V~\cite{yao2024minicpm}}  & \multirow{2}{*}{\textit{Adversarial-A}}           & Regular~\cite{yao2024minicpm}           &11.15 &36.16 &11.32 &17.24                                         \\&                               & AGLA               &\textbf{14.99} &36.40 &15.03 &\textbf{21.28}  \\
                          \cline{2-7} 
                          & \multirow{2}{*}{\textit{Adversarial-B}}     & Regular~\cite{yao2024minicpm}           &11.21 &30.73 &11.38 &16.61  \\
                                    &                               & AGLA               &\textbf{15.82} &33.23 &15.87 &\textbf{21.48}   \\ 
                        \cline{2-7}
                          & \multirow{2}{*}{\textit{Heterogeneous}} &  Regular~\cite{yao2024minicpm}           &3.22 &7.69 &3.41 &4.73  \\
                                   &                               & AGLA               &\textbf{4.35} &8.40 &4.39 &\textbf{5.77}  \\
                        \cline{2-7}
                          & \multirow{2}{*}{\textit{Homogenous}} &  Regular~\cite{yao2024minicpm}           &18.32 &49.43 &18.79 &27.23  \\
                                   &                               & AGLA               &\textbf{25.39} &64.98 &25.39 &\textbf{36.51}  \\
                        \cline{2-7}
                          & \multirow{2}{*}{\textit{Mixed}} &  Regular~\cite{yao2024minicpm}           &8.35 &21.79 &8.59 &12.32  \\
                                   &                               & AGLA               &\textbf{10.70} &25.95 &10.73 &\textbf{15.19}  \\

                            \hline\hline
\end{tabular}
}
\vspace{-2mm}
\label{tab:rope}
\end{table*}

\noindent \textbf{Implementation Details.}
We evaluate the effectiveness of AGLA on several state-of-the-art LVLMs including LLaVA-1.5~(7B and 13B)~\cite{liu2023visual}, InstructBLIP~(7B and 13B)~\cite{dai2023instructblip}, QWen-VL~(7B)~\cite{bai2023qwen}, MiniCPM-V~(2.4B)~\cite{yao2024minicpm}. Multinomial sampling is employed as the decoding strategy (denoted as \textbf{Regular}). We also compare AGLA with three state-of-the-art decoding methods including OPERA~\cite{huang2023opera}, DOLA~\cite{chuang2023dola} and VCD~\cite{vcd}. We use BLIP-ITM~\cite{li2022blip} as the matching model and report average results over 3 runs for each benchmark.
For compared models, we follow the suggested settings in their respective papers and released codes to ensure fair comparisons. More implementation details are provided in Appendix~\ref{app_details}.

\subsection{Experimental Results}
\noindent\textbf{Experiments on POPE.} 
Table~\ref{tab:average} shows experimental results on the three POPE subsets~\cite{li2023evaluating} with LLaVA-1.5~(7B) and InstructBLIP~(7B). It is obvious that AGLA consistently outperforms the \textit{Regular} decoding strategy by substantial margins (average improvements of 5.5\% in accuracy and 5.1\% in F1 score) across all LVLMs and settings. Additionally, AGLA surpasses state-of-the-art decoding methods as well, demonstrating its effectiveness in mitigating object hallucinations.
Furthermore, the performance improvement under the adversarial setting confirms that our model effectively resolves \textit{attention deficiency} and mitigates object hallucinations caused by object association.
Detailed results and additional results for other models are provided in Appendix~\ref{app_pope},~\ref{app_pope_more} and~\ref{app_pope_scale}.

\noindent\textbf{Multiple Object Hallucination Mitigation.} 
Table~\ref{tab:rope} shows experimental results on the five ROPE subsets~\cite{chen2024multiobject} with LLaVA-1.5 (7B) and MiniCPM-V (2.4B). Notably, AGLA consistently outperforms the \textit{Regular} decoding strategy by significant margins across all LVLMs and configurations, demonstrating the effectiveness of our model in mitigating multi-object hallucinations.
The improved performance also confirms that the proposed IPM module can effectively handle queries containing multiple objects and select query-related image areas, demonstrating its effectiveness in complex real-world applications. Qualitative examples are provided in Appendix~\ref{app_multiobject}.

\begin{table}[t]
\centering
\vspace{1mm}
\caption{Results of CHAIR hallucination evaluation for the open-ended caption generation task.}
\vspace{-2mm}
\renewcommand\arraystretch{1}
\scalebox{0.9}
{%
\setlength{\tabcolsep}{6.5pt}
\begin{tabular}{clccc}
\hline\hline
Model & Decoding                        & C$_{S}$ $\downarrow$ & C$_{I}$ $\downarrow$ & Recall $\uparrow$ \\ \hline\hline
\multirow{5}{*}{LLaVA-1.5~\cite{liu2023visual}}     & Regular~\cite{vcd}  & 51.0 & 15.2 & 75.2 \\
& DOLA~\cite{chuang2023dola}     & 57.0 & 15.9 & 78.2 \\
& VCD~\cite{vcd}      & 51.0 & 14.9 & 77.2 \\
& OPERA~\cite{huang2023opera}    & 47.0 & 14.6 & 78.5 \\
& \textbf{AGLA}      & \textbf{43.0} & \textbf{14.1} & \textbf{78.9} \\ \hline\hline
\multirow{5}{*}{InstructBLIP~\cite{dai2023instructblip}}     
& Regular~\cite{vcd}  & 54.0 & 18.1 & 71.1 \\
& DOLA~\cite{chuang2023dola}     & 60.0 & 20.1 & 71.5 \\
& VCD~\cite{vcd}      & 57.0 & 17.0 & 72.1 \\
& OPERA~\cite{huang2023opera}    & 54.0 & 12.8 & 69.8 \\
& \textbf{AGLA}      & \textbf{49.0} & \textbf{12.1} & \textbf{72.5} \\ \hline\hline
\end{tabular}
}
\label{tab:chair}
\vspace{-4mm}
\end{table}

\begin{figure*}[t]
\vspace{-8mm}
\begin{subfigure}{0.49\textwidth}
  \centering
  \includegraphics[width=\linewidth]{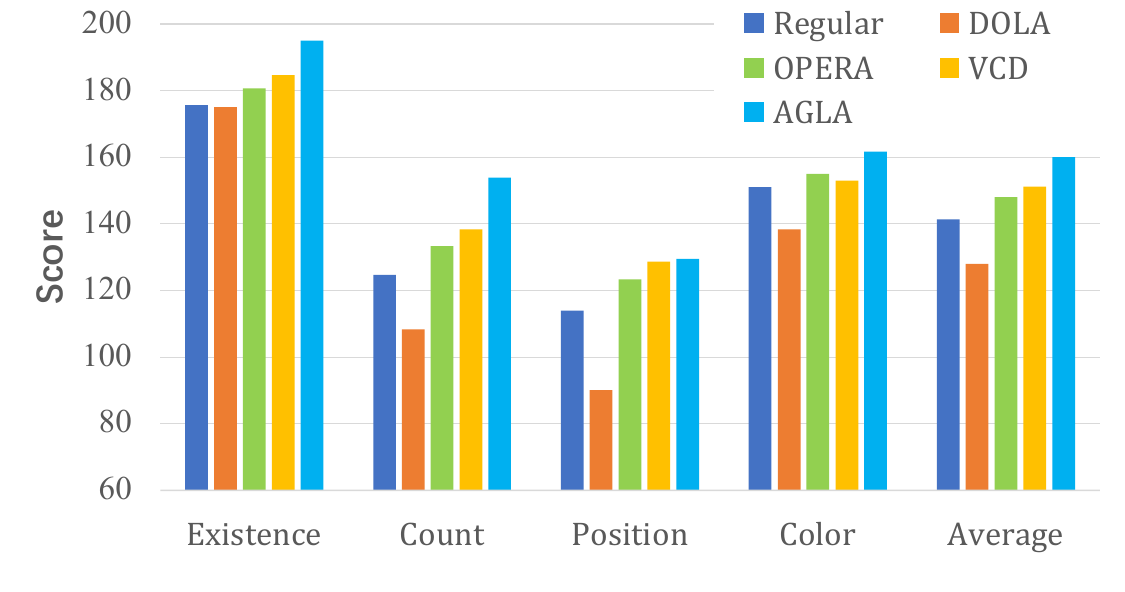}
\end{subfigure}
\vspace{-4mm}
\begin{subfigure}{0.49\textwidth}
  \centering
  \includegraphics[width=\linewidth]{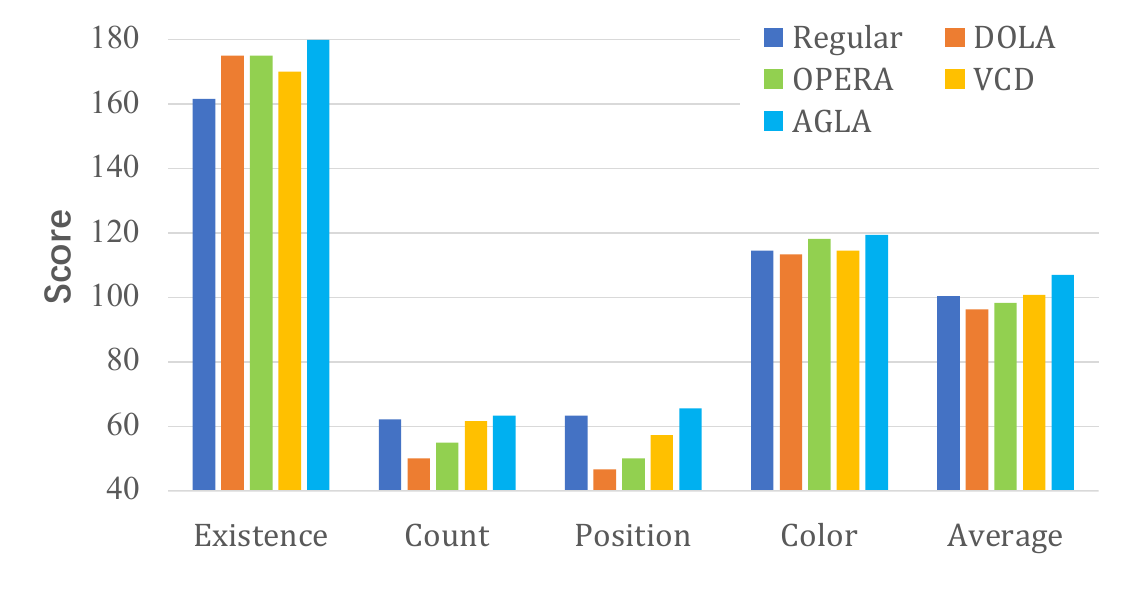}
\end{subfigure}
\vspace{-2mm}
\caption{Hallucination mitigation over the dataset MME with LLaVA-1.5 (Left) and InstructBLIP (Right).}
\vspace{-2mm}
\label{fig:mme}
\end{figure*}

\noindent\textbf{Experiments on MME.} 
We conduct comprehensive experiments on the MME hallucination subset, which includes four types of hallucinations that assess the overall ability of LVLMs from the object level (\ie, Existence), attribute level (\ie, Count and Color) and relation level (\ie, Position). As illustrated in Fig.~\ref{fig:mme}, the proposed model consistently outperforms both \textit{Regular} and SOTA decoding strategies across all hallucination categories and LVLMs. This underscores AGLA's effectiveness in addressing a broader range of multimodal hallucinations beyond object hallucinations.
To further evaluate the impact of AGLA on the perception capabilities of LVLMs, we conduct experiments on MME perception subsets in Appendix~\ref{app_mme}. The improved performance further validates the effectiveness of AGLA in enhancing the general perception capabilities of LVLMs.

\noindent\textbf{Experiments on CHAIR.} 
Beyond the ``Yes-or-No'' discriminative evaluations on the POPE, ROPE, and MME datasets, we also validate our model on open-ended caption generation using the CHAIR benchmark~\cite{chair}. Experiments in Table~\ref{tab:chair} show that the proposed AGLA  achieves consistent improvements over the compared methods. 
Specifically, AGLA effectively reduces object hallucinations in generated captions, as evidenced by lower CHAIR$_S$ and CHAIR$_I$ scores. In addition, AGLA enhances the detailedness of the generated captions, as indicated by higher Recall scores. In summary, AGLA achieves a good balance between accuracy and detailedness in open-ended caption generation, benefiting from the integration of the discriminative local features and the generative global features as described in Section~\ref{sec42}.

\begin{table}[t]
\centering
\vspace{0mm}
\caption{Results of GPT-4-aided evaluation on the LLaVA-Bench-Wild dataset. Both metrics are scored on a scale of 10.}
\vspace{-1mm}
\renewcommand\arraystretch{1.05}
\scalebox{0.88}
{
\setlength{\tabcolsep}{9.8pt}
\begin{tabular}{clcc}

\hline\hline
Model & Decoding                        & Acc. $\uparrow$ & Detail. $\uparrow$  \\ 
\hline\hline
\multirow{5}{*}{LLaVA-1.5~\cite{liu2023visual}}     
& Regular~\cite{vcd}  & 2.61 & 3.65  \\
& VCD~\cite{vcd}      & 3.00 & 3.83 \\
& DOLA~\cite{chuang2023dola}      & 3.51 & 4.22 \\
& OPERA~\cite{huang2023opera}    & 3.68 & 4.29 \\
& \textbf{AGLA}      & \textbf{3.83} & \textbf{4.39} \\ \hline\hline
\multirow{5}{*}{InstructBLIP~\cite{dai2023instructblip}}     
& Regular~\cite{vcd}  & 2.82 & 3.36 \\ 
& DOLA~\cite{chuang2023dola}     & 4.04 & 4.33 \\
& VCD~\cite{vcd}      & 4.50 & 4.55 \\
& OPERA~\cite{huang2023opera}    & 4.56 & 4.48 \\
& \textbf{AGLA}      & \textbf{4.59} & \textbf{4.59}  \\ 
\hline\hline
\end{tabular}
}
\vspace{-2mm}
\label{tab:gpt4}
\end{table}

\noindent \textbf{Experiments on LLaVA-Bench-Wild.} 
Following previous studies~\cite{huang2023opera,vcd}, we also assess the overall quality of the generated captions on the LLaVA-Bench-Wild dataset~\cite{liu2023visual} with GPT-4. 
As Table~\ref{tab:gpt4} shows, the proposed AGLA outperforms other methods across all evaluation metrics. Specifically, the improved accuracy indicates that AGLA effectively mitigates hallucinations in the generated captions by incorporating local and global image attention. And the enhancement in detailedness highlights AGLA's ability to improve the global perceptual capabilities of LVLMs, enabling the generation of more comprehensive captions.

\begin{table}[t]
\centering
\vspace{0mm}
\caption{Ablation study with different model variants and masking strategies on POPE-COCO under the popular setting.}
\vspace{-1mm}
\renewcommand\arraystretch{1.07}
\scalebox{0.92}
{
\setlength{\tabcolsep}{11.5pt}
\begin{tabular}{lcc}
\hline\hline
Model Variants                       & Accuracy $\uparrow$  & F1 Score $\uparrow$  \\ \hline\hline
\textbf{AGLA}     & \textbf{86.12} & \textbf{84.71}  \\
\hline\hline
\multicolumn{3}{c}{\textit{\textbf{Models}}} \\\hline
w/o Truncation       & 85.66 & 84.42 \\
w/o Adaptive Ratio   & 84.83 & 82.94  \\
w/o Assembly         & 83.53 & 82.14 \\
Regular~\cite{vcd}   & 81.88 & 80.06  \\
\hline\hline
\multicolumn{3}{c}{\textit{\textbf{Masking Strategy}}} \\\hline
Patch                & 85.77 & 84.05 \\
Soft                 & 85.51 & 83.93 \\
Feature              & 85.27 & 83.73 \\
Random               & 83.56 & 82.40 \\
\hline\hline
\end{tabular}
}
\label{tab:ablation}
\vspace{-2mm}
\end{table}

\begin{figure*}[t]
\centering
\vspace{-8mm}
\includegraphics[width=0.98\linewidth]{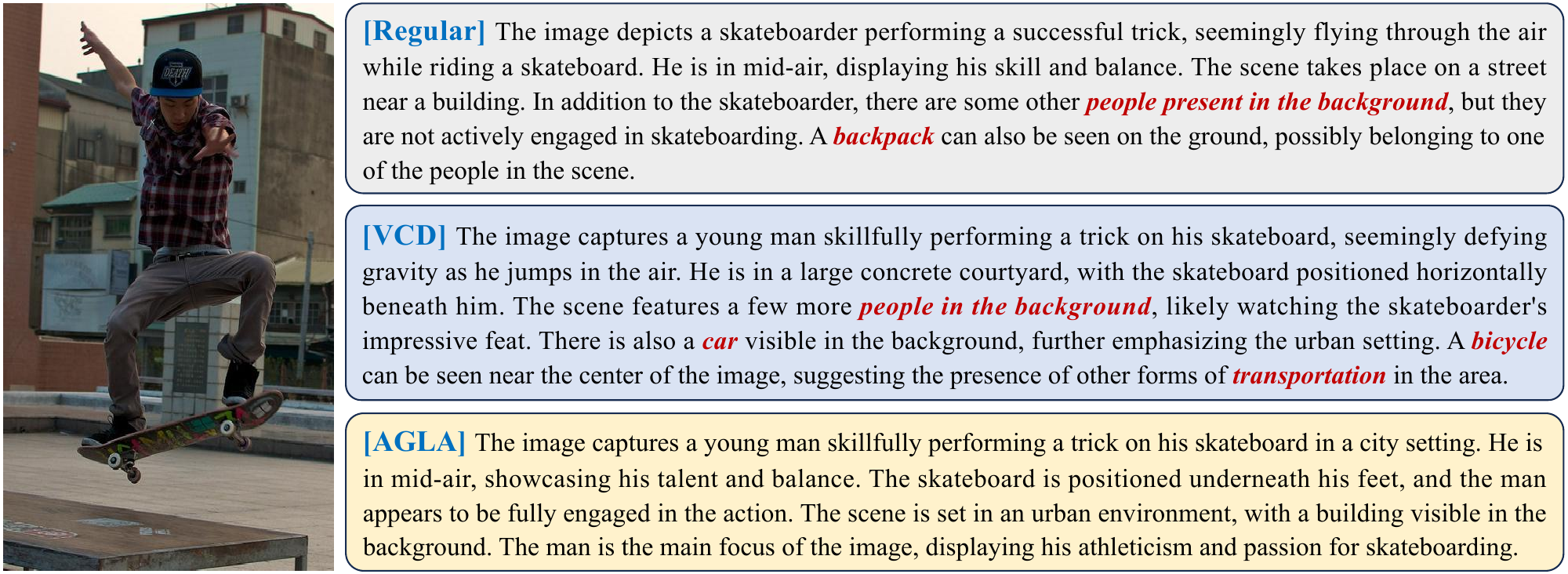}
\vspace{-1.5mm}
\caption{Captions generated by different decoding methods. Hallucinated contents are highlighted in \textcolor{red}{\textbf{\textit{red}}}.} 
\label{fig5}
\vspace{2.5mm}
\end{figure*}

\begin{table*}[tp]
\vspace{-2.5mm}
\caption{An ablation study of different decoding strategies. Regular means sampling from the original distribution.}
\vspace{-2mm}
\centering
\scalebox{0.85}
{
\setlength{\tabcolsep}{6.7pt}
\begin{tabular}{lccccc|lccccc}
\hline\hline
\multicolumn{1}{l}{Decoding}    &  Model    & Accuracy & Precision & Recall & F1  & \multicolumn{1}{l}{Decoding}          &  Model    & Accuracy & Precision & Recall & F1  \\ 
\hline\hline
\multirow{2}{*}{Top P}       
& Regular  & 80.13    & 84.45     & 73.87  & 78.81 & \multirow{2}{*}{Top P + Temp.} & Regular  & 81.50    & 86.49     & 74.67  & 80.14 \\ & AGLA & \textbf{84.23}    & 89.84     & 77.20  & \textbf{83.04} &&  AGLA & \textbf{84.50}    & 90.15     & 77.47  & \textbf{83.33} \\
\midrule
\multirow{2}{*}{Top K}         
& Regular  & 78.50    & 82.07     & 72.93  & 77.23 & \multirow{2}{*}{Top K + Temp.} & Regular  & 80.60    & 85.20     & 74.07  & 79.24 \\
& AGLA & \textbf{83.80}    & 89.24    & 76.87  & \textbf{82.59} && AGLA & \textbf{84.33}    & 89.92     & 77.33  & \textbf{83.15} \\
\midrule
\multirow{2}{*}{Temp.} & Regular & 81.77    &  86.91    & 74.80   & 80.40 
&\multirow{2}{*}{Greedy} & Regular  & 83.63    & 89.44     & 76.27   & 82.33 \\ &AGLA & \textbf{84.33}    & 89.92     & 77.33  & \textbf{83.15}  && AGLA & \textbf{85.00}   & 90.70  & 78.00  & \textbf{83.87} \\

\hline\hline
\end{tabular}
}
\label{tab:ablation4}
\vspace{-4mm}
\end{table*}

\subsection{Ablation Study}
We conduct two sets of ablation studies to evaluate the effectiveness of each design in our model. Table~\ref{tab:ablation} shows experimental results on the POPE-COCO with LLaVA-1.5. Specifically, we first test three AGLA variants including: \textit{(1) Truncation Removal:} that omits truncation as defined in Eq.~\ref{constraint}; \textit{(2) Fixed Masking Ratio:} that replaces the adaptive masking by a fixed masking ratio; and \textit{(3) Assembly Removal:} that removes the original inputs image and uses the augmented view only. It can be observed that \textit{Truncation Removal} leads to performance drop, indicating that adaptive plausibility constraints are crucial for preserving essential global information and preventing the calibrated distribution from deviating excessively from the original distribution. \textit{Fixed Masking Ratio} leads to a clear performance drop as well, largely because a fixed ratio cannot accurately capture the model's matching confidence for different image-prompt pairs. Further, the large performance drop by \textit{Assembly Removal} demonstrates the critical role of the original image distribution which helps provide global image information. Nevertheless, we can observe that all three AGLA variants outperform the regular decoding (Regular), demonstrating the overall effectiveness of our model.

In addition, we examine how ALGA is affected by different masking strategies while generating the augmented image view. As Table~\ref{tab:ablation} shows, Patch-level Masking (Patch) degrades the performance slightly due to its nature of coarse-grained masking. Soft Masking (Soft) using relevance scores as weights leads to more performance drops as important information could be masked due to the highly variable relevance values. Feature-level Masking (Feature) leads to further performance drop because self-attention may cause information leakage. On the other end, all masking strategies outperform random masking (Random), highlighting the necessity of the matching model. Details of different masking strategies are provided in Appendix~\ref{app_mask}.

\subsection{Discussion}

\noindent\textbf{Effect of Different Decoding Strategies.}
Besides direct sampling, we also present an ablation study with various decoding strategies on the POPE-COCO dataset with the adversarial setting using LLaVA-1.5~\cite{liu2023visual}. The experiment includes six kinds of additional decoding strategies: Top P sampling~\cite{holtzman2019curious} ($p=0.7$), Top K sampling~\cite{fan2018hierarchical} ($k=50$), Temperature sampling~\cite{ackley1985learning} ($t=0.5$), Top P sampling with temperature ($p=0.7$ and $t=0.5$), Top K sampling with temperature ($k=50$ and $t=0.5$) and Greedy decoding~\cite{devore1996some} ($\alpha=1$ and $\beta=0.1$). The results in Table~\ref{tab:ablation4} show that \textit{AGLA} consistently contributes to mitigating hallucinations, irrespective of the decoding strategies adopted.

\noindent\textbf{Qualitative Results.} 
We present qualitative illustrations of different decoding methods in Fig.~\ref{fig5}. We can see that \textit{Regular} decoding~\cite{liu2023visual} and VCD~\cite{vcd} often produce object hallucinations, especially when the generated captions become longer. These issues are largely attributed to \textit{attention deficiency} and association bias. In contrast, AGLA assembles local and global features to mitigate \textit{attention deficiency} and masks out distractions to alleviate association bias, suppressing hallucinations greatly in the generated text description. 
More examples are provided in the Appendix~\ref{app_example}.
\vspace{-3mm}

\section{Conclusion}
In this paper, we investigate the underlying reasons for object hallucinations in LVLMs, identifying \textit{attention deficiency} as one of the root causes. To address this, we propose \textit{Assembly of Global and Local Attention (AGLA)}, a training-free, plug-and-play approach that mitigates object hallucinations by combining local and global model attention to capture both discriminative and generative image features.
Our approach features an image-prompt matching scheme that captures prompt-relevant local attention toward images, resulting in an augmented view of the input image where prompt-irrelevant distractions are masked. We then integrate generative global features from original images with discriminative local features from augmented images to derive a calibrated decoding distribution through logit fusion, thereby enhancing the generation and perception capabilities of LVLMs.
Experiments on various discriminative and generative benchmarks demonstrate that our model significantly reduces object hallucinations and improves the general perception capabilities of LVLMs.

\section{Acknowledgments}
This work was supported by the National Science and Technology Major Project (2022ZD0117102), National Natural Science Foundation of China (62177038, 62293551, 62277042, 62377038). Project of China Knowledge Centre for Engineering Science and Technology, ``LENOVO-XJTU" Intelligent Industry Joint Laboratory Project, Program of China Scholarship Council (202406280146).

{
    \small
    \bibliographystyle{ieeenat_fullname}
    \bibliography{main}
}

\clearpage
\setcounter{page}{1}
\maketitlesupplementary

\setcounter{table}{0}
\setcounter{figure}{0}
\setcounter{section}{0}
\setcounter{equation}{0}
\renewcommand{\thetable}{A\arabic{table}}
\renewcommand{\thefigure}{A\arabic{figure}}
\renewcommand{\thesection}{A\arabic{section}}
\renewcommand{\theequation}{A\arabic{equation}}

\section{Limitation and Future Work}
\label{limitation}
Despite the superb performance in mitigating hallucinations and enhancing the general perception capabilities of LVLMs, our work could be improved in several aspects. First, we conducted experiments on the most widely used LVLMs due to resource constraints. It will be useful to evaluate our model on larger LVLMs such as LLaVA 34B and Flamingo 70B \cite{alayrac2022flamingo}. In addition, this work focuses on text and image data. It could be extended to data from other modalities such as videos. We will examine these problems in our future work.

\section{Evaluation Metrics}
\label{app_metric}
The details of evaluation metrics for different datasets are listed below.

\noindent \textbf{POPE.} Since the output of the model for the POPE dataset~\cite{li2023evaluating} is limited to two types (Yes or No), it is convenient to measure the model performance with binary classification metrics: Accuracy, Precision, Recall, and F1 score.

\noindent \textbf{ROPE.} We utilize the official code of ROPE~\cite{chen2024multiobject} to measure the multi-object hallucination mitigation performance of different models. Similar to POPE, the evaluation metrics include Accuracy, Precision, Recall, and F1 score.

\noindent \textbf{MME.} Similar to the POPE dataset, the MME dataset~\cite{fu2023mme} contains only two types of answers (\ie, Yes or No). Following the setting in their original paper, we use the sum of accuracy and accuracy+ as the final score, where accuracy is calculated based on each question and accuracy+ is calculated based on each image where both of the two questions need to be answered correctly. So accuracy+ is a stricter measurement that can better reflect the comprehensive understanding degree of the model.

\noindent \textbf{CHAIR.} Different from discriminative datasets like POPE and MME, CHAIR~\cite{chair} (\ie, Caption Hallucination Assessment with Image Relevance) is a framework that quantifies object hallucinations for generative datasets. The sentence-level score CHAIR$_{S}$ represents the proportion of generated captions that contain hallucinations, while the instance-level score CHAIR$_{I}$ denotes the proportion of hallucinated objects relative to all mentioned objects in the generated captions. In addition, we also evaluate the semantic detailedness of generated captions with the metric Recall. CHAIR$_{S}$, CHAIR$_{I}$, and Recall are computed as follows.
\begin{equation}
    C_{S} = \frac{|\{\text{\scriptsize Captions with hallucinated objects}\}|}{|\{\text{\scriptsize All captions}\}|}
\end{equation}
\begin{equation}
    C_{I} = \frac{|\{\text{\scriptsize Hallucinated objects}\}|}{|\{\text{\scriptsize All mentioned objects}\}|}
\end{equation}

\begin{equation}
    Recall = \frac{|\{\text{\scriptsize Accurate objects}\}|}{|\{\text{\scriptsize Ground-truth objects}\}|}
\end{equation}

\noindent \textbf{LLaVA-Bench-Wild.} LLaVA-Bench-Wild~\cite{liu2023visual} contains 24 images with 60 questions to assess the capability of LVLMs in tackling challenging tasks and their adaptability to new domains. Following previous works~\cite{huang2023opera,vcd}, we use GPT-4 to evaluate the accuracy and detailedness of the generated captions. Specifically, the metric Accuracy measures the captions' alignment with the image content and the metric Detailedness gauges the richness of details in the generated captions. The specific configurations used to prompt GPT-4 are listed in Table~\ref{tab:prompt}.

\section{Experimental Settings}
\label{app_details}
For the POPE~\cite{li2023evaluating} and ROPE~\cite{chen2024multiobject} datasets, $\alpha$ is set to 2 and $\beta$ is set to 0.5.
For the MME~\cite{fu2023mme} dataset, we set $\alpha$ to 2 and $\beta$ to 0.5 for LLaVA-1.5, while setting $\alpha$ and $\beta$ to 0.1 for InstructBLIP. 
For the CHAIR~\cite{chair} and LLaVA-Bench-Wild~\cite{liu2023visual} datasets, $\alpha$ is set to 2 and $\beta$ is set to 0.5.
All 7B model experiments are conducted on a single RTX 3090 GPU, and all 13B model experiments are conducted on two RTX 3090 GPUs.

\begin{table*}[tp]
\caption{Results on the full POPE dataset with LLaVA-1.5 7B~\cite{liu2023visual} and InstructBLIP 7B~\cite{dai2023instructblip}.}
\centering
{%
\begin{tabular}{cllllll|l}
\hline
\textbf{Dataset}          & \textbf{Setting}                         & \textbf{Model}                & \textbf{Decoding} & Accuracy $\uparrow$ & Precision & Recall & F1 Score $\uparrow$  \\ \hline
\multirow{12}{*}{COCO}  & \multirow{4}{*}{\textit{Random}}      & \multirow{2}{*}{LLaVA-1.5~\cite{liu2023visual}}     & Regular           &83.29 &92.13 &72.80 &81.33                                         \\
                          &                                       &                               & AGLA               &\textbf{87.46} &97.52 &76.87 &\textbf{85.97}  \\
                          &                                       & \multirow{2}{*}{InstructBLIP~\cite{dai2023instructblip}} & Regular           &80.71 &81.67 &79.19 &80.41  \\
                          &                                       &                               & AGLA               &\textbf{87.63} &93.88 &80.51 &\textbf{86.68}  \\ \cline{2-8} 
                          & \multirow{4}{*}{\textit{Popular}}     & \multirow{2}{*}{LLaVA-1.5~\cite{liu2023visual}}     & Regular           &81.88 &88.93 &72.80 &80.06  \\
                          &                                       &                               & AGLA               &\textbf{86.12} &94.33 &76.87 &\textbf{84.71}   \\

                          &                                       & \multirow{2}{*}{InstructBLIP~\cite{dai2023instructblip}} & Regular           &78.22 &77.87 &78.85 &78.36  \\
                          &                                       &                               & AGLA               &\textbf{84.63} &87.75 &80.51 &\textbf{83.97}  \\ \cline{2-8} 
                          & \multirow{4}{*}{\textit{Adversarial}} & \multirow{2}{*}{LLaVA-1.5~\cite{liu2023visual}}     & Regular           &78.96 &83.06 &72.75 &77.57  \\
                          &                                       &                               & AGLA               &\textbf{83.87} & 89.48  &76.76 &\textbf{82.63}  \\
                          &                                       & \multirow{2}{*}{InstructBLIP~\cite{dai2023instructblip}} & Regular           &75.84 &74.30 &79.03 &76.59  \\
                          &                                       &                               & AGLA               &\textbf{81.90} &82.76 &80.58 &\textbf{81.66}  \\ \hline
\multirow{12}{*}{VQA} & \multirow{4}{*}{\textit{Random}}      & \multirow{2}{*}{LLaVA-1.5~\cite{liu2023visual}}     & Regular           &83.45 &87.24 &78.36 &82.56  \\
                          &                                       &                               & AGLA               &\textbf{89.28} &93.18 &84.76 &\textbf{88.77}  \\
                          &                                       & \multirow{2}{*}{InstructBLIP~\cite{dai2023instructblip}} & Regular           &80.91 &77.97 &86.16 &81.86  \\
                          &                                       &                               & AGLA               &\textbf{87.80} &86.76 &89.22 &\textbf{87.97}  \\ \cline{2-8} 
                          & \multirow{4}{*}{\textit{Popular}}     & \multirow{2}{*}{LLaVA-1.5~\cite{liu2023visual}}     & Regular           &79.90 &80.85 &78.36 &79.59  \\
                          &                                       &                               & AGLA               &\textbf{85.63} &86.27 &84.76 &\textbf{85.51}  \\
  
                          &                                       & \multirow{2}{*}{InstructBLIP~\cite{dai2023instructblip}} & Regular           &76.19 &72.16 &85.28 &78.17  \\
                          &                                       &                               & AGLA               &\textbf{82.27} &78.33 &89.22 &\textbf{83.42}  \\ \cline{2-8} 
                          & \multirow{4}{*}{\textit{Adversarial}} & \multirow{2}{*}{LLaVA-1.5~\cite{liu2023visual}}     & Regular           &74.04 &72.08 &78.49 &75.15  \\
                          &                                       &                               & AGLA               &\textbf{78.85} &75.83 &84.71 &\textbf{80.03}  \\
   
                          &                                       & \multirow{2}{*}{InstructBLIP~\cite{dai2023instructblip}} & Regular           &70.71 &65.91 &85.83 &75.56  \\
                          &                                       &                               & AGLA               &\textbf{74.79} &69.33 &88.91 &\textbf{77.91}  \\ \hline
\multirow{12}{*}{GQA}     & \multirow{4}{*}{\textit{Random}}      & \multirow{2}{*}{LLaVA-1.5~\cite{liu2023visual}}     & Regular           &83.73 &87.16 &79.12 &82.95  \\
                          &                                       &                               & AGLA               &\textbf{88.89} &92.53 &84.60 &\textbf{88.39}  \\
                          
                          &                                       & \multirow{2}{*}{InstructBLIP~\cite{dai2023instructblip}} & Regular           &79.65 &77.14 &84.29 &80.56  \\
                          &                                       &                               & AGLA               &\textbf{86.46} &85.84 &87.31 &\textbf{86.57}  \\ \cline{2-8} 
                          & \multirow{4}{*}{\textit{Popular}}     & \multirow{2}{*}{LLaVA-1.5~\cite{liu2023visual}}     & Regular           &78.17 &77.64 &79.12 &78.37  \\
                          &                                       &                               & AGLA               &\textbf{83.67} &83.05 &84.60 &\textbf{83.82}  \\

                          &                                       & \multirow{2}{*}{InstructBLIP~\cite{dai2023instructblip}} & Regular           &73.87 &69.63 &84.69 &76.42  \\
                          &                                       &                               & AGLA               &\textbf{78.67} &74.44 &87.31 &\textbf{80.36}  \\ \cline{2-8} 
                          & \multirow{4}{*}{\textit{Adversarial}} & \multirow{2}{*}{LLaVA-1.5~\cite{liu2023visual}}     & Regular           &75.08 &73.19 &79.16 &76.06  \\
                          &                                       &                               & AGLA               &\textbf{80.66} &78.30 &84.82 &\textbf{81.43}  \\

                          &                                       & \multirow{2}{*}{InstructBLIP~\cite{dai2023instructblip}} & Regular           &70.56 &66.12 &84.33 &74.12  \\
                          &                                       &                               & AGLA               &\textbf{75.18} &70.19 &87.53 &\textbf{77.91}  \\ \hline
\end{tabular}
}
\label{tab:fullpope}
\end{table*}

\section{Different Masking Strategies}
\label{app_mask}
In our paper, we mask the image at the pixel level by setting the RGB value of the masking pixel to 0. ``Patch-level" masking means masking the image patches based on the activation value of the patch regions. ``Soft" masking involves multiplying the GradCAM activation value by the original pixel values instead of setting the masking pixels to 0. ``Feature-level" masking means masking the image features extracted by the vision encoder rather than masking the input image. ``Random" masking means randomly masking the pixels of the input images. The results of different masking strategies are listed in Table~\ref{tab:ablation}.

\section{Detailed Results on POPE}
\label{app_pope}
The results of LLaVA-1.5 7B~\cite{liu2023visual} and InsturctBLIP 7B~\cite{dai2023instructblip} on the full POPE dataset~\cite{li2023evaluating} with three subsets (\ie, MSCOCO~\cite{lin2014microsoft}, A-OKVQA~\cite{schwenk2022okvqa} and GQA~\cite{hudson2019gqa}) and three negative sample settings (\ie, Random, Popular and Adversarial) are listed in Table~\ref{tab:fullpope}. From the table, we can see that the proposed decoding strategy \textit{AGLA} consistently outperforms the regular decoding strategy by large margins (average 5.5\% accuracy and 5.1\% F1 improvement) on all LVLMs. The improved performance on popular and adversarial (\ie, co-occurrence) settings of different subsets validates that our model can better mitigate the statistical bias of LVLMs by addressing the \textit{attention deficiency} issue.

\section{Results on POPE with More LVLMs}
\label{app_pope_more}
In addition to LLaVA-1.5~\cite{liu2023visual} and InsturctBLIP~\cite{dai2023instructblip}, we perform experiments with more kinds of LVLMs, such as Qwen-VL~\cite{bai2023qwen} and MiniCPM-V~\cite{yao2024minicpm}. The results are shown in Table~\ref{tab:fullpopeQWenVL} and~\ref{tab:minicpm}, respectively. From the tables, we can see that the proposed \textit{AGLA} consistently outperforms the regular decoding strategy by large margins on all evaluation metrics and LVLMs, which can show the generalizability of our method towards different kinds of LVLMs.

\begin{table*}[tp]
\caption{Results on the full POPE dataset with QWen-VL~\cite{bai2023qwen}.}
\centering
\scalebox{1.1}
{
\begin{tabular}{cllccc|c}
\hline
\textbf{Dataset}          & \textbf{Setting}                  & \textbf{Decoding} & Accuracy $\uparrow$  & Precision & Recall & F1 Score $\uparrow$   \\ \hline
\multirow{6}{*}{COCO}  & \multirow{2}{*}{\textit{Random}}           & Regular           &83.17 &96.80 &68.60 &80.30                                         \\&                               & AGLA               &\textbf{84.60} &98.23 &70.47 &\textbf{82.07}  \\
                          \cline{2-7} 
                          & \multirow{2}{*}{\textit{Popular}}     & Regular           &82.76 &95.63 &68.66 &79.94  \\
                                    &                               & AGLA               &\textbf{84.40} &97.16 &70.87 &\textbf{81.95}   \\

                        \cline{2-7}
                          & \multirow{2}{*}{\textit{Adversarial}} &  Regular           &82.00 &92.47 &69.67 &79.47  \\
                                   &                               & AGLA               &\textbf{82.70} &93.22 &70.53 &\textbf{80.30}  \\

                            \hline
\multirow{6}{*}{VQA}  & \multirow{2}{*}{\textit{Random}}           & Regular           &84.53 &93.31 &74.40 &82.79                                         \\ &                               & AGLA               &\textbf{86.77} &95.02 &77.60 &\textbf{85.43}  \\

                          \cline{2-7} 
                          & \multirow{2}{*}{\textit{Popular}}     &  Regular           &84.43 &92.44 &75.00 &82.81  \\
                                    &                               & AGLA               &\textbf{86.30} &94.09 &77.47 &\textbf{84.97}   \\ 

                        \cline{2-7}
                          & \multirow{2}{*}{\textit{Adversarial}} &  Regular           &78.20 &80.61 &74.27 &77.31  \\
 &                               & AGLA               &\textbf{80.73} &82.79 &77.60 &\textbf{80.11}  \\
            
                            \hline
\multirow{6}{*}{GQA}  & \multirow{2}{*}{\textit{Random}}           & Regular           &81.20 &90.48 &69.73 &78.76                                         \\
             &                               & AGLA               &\textbf{83.90} &93.05 &73.26 &\textbf{81.98}  \\ 
                          \cline{2-7} 
                          & \multirow{2}{*}{\textit{Popular}}     & Regular           &78.23 &82.90 &71.13 &76.56  \\
               &                               & AGLA               &\textbf{80.80} &85.70 &73.93 &\textbf{79.38}   \\ 
                        \cline{2-7}
                          & \multirow{2}{*}{\textit{Adversarial}} &  Regular           &75.63 &79.06 &69.73 &74.10  \\
            &                               & AGLA               &\textbf{78.73} &82.16 &73.40 &\textbf{77.53}  \\
                            \hline
\end{tabular}
}
\label{tab:fullpopeQWenVL}
\end{table*}

\begin{table*}[tp]
\caption{Results on the full POPE dataset with MiniCPM-V~\cite{yao2024minicpm}.}
\centering
\scalebox{1.1}
{
\begin{tabular}{cllccc|c}
\hline
\textbf{Dataset}          & \textbf{Setting}                  & \textbf{Decoding} & Accuracy $\uparrow$  & Precision & Recall & F1 Score $\uparrow$   \\ \hline
\multirow{6}{*}{COCO}  & \multirow{2}{*}{\textit{Random}}           & Regular           &86.00 &87.09 &84.53 &85.79                                         \\&                               & AGLA               &\textbf{88.90} &92.50 &84.67 &\textbf{88.41}  \\
                          \cline{2-7} 
                          & \multirow{2}{*}{\textit{Popular}}     & Regular           &84.00 &83.20 &85.20 &84.19  \\
                                    &                               & AGLA               &\textbf{85.76} &86.62 &84.60 &\textbf{85.60}   \\

                        \cline{2-7}
                          & \multirow{2}{*}{\textit{Adversarial}} &  Regular           &80.27 &77.99 &84.33 &81.03  \\
                                   &                               & AGLA               &\textbf{82.26} &80.86 &84.53 &\textbf{82.66}  \\

                            \hline
\multirow{6}{*}{VQA}  & \multirow{2}{*}{\textit{Random}}           & Regular           &84.70 &81.56 &89.66 &85.43                                         \\ &                               & AGLA               &\textbf{88.03} &85.36 &91.80 &\textbf{88.47}  \\

                          \cline{2-7} 
                          & \multirow{2}{*}{\textit{Popular}}     &  Regular           &79.30 &74.40 &89.33 &81.19  \\
                                    &                               & AGLA               &\textbf{82.30} &77.08 &91.93 &\textbf{83.85}   \\ 

                        \cline{2-7}
                          & \multirow{2}{*}{\textit{Adversarial}} &  Regular           &72.33 &66.43 &90.26 &76.54  \\
 &                               & AGLA               &\textbf{73.80} &67.53 &91.66 &\textbf{77.77}  \\
            
                            \hline
\multirow{6}{*}{GQA}  & \multirow{2}{*}{\textit{Random}}           & Regular           &84.63 &81.89 &88.93 &85.27                                         \\
             &                               & AGLA               &\textbf{87.06} &84.41 &90.93 &\textbf{87.54}  \\ 
                          \cline{2-7} 
                          & \multirow{2}{*}{\textit{Popular}}     & Regular           &74.93 &69.33 &89.40 &78.10  \\
               &                               & AGLA               &\textbf{77.26} &71.52 &90.60 &\textbf{79.94}   \\ 
                        \cline{2-7}
                          & \multirow{2}{*}{\textit{Adversarial}} &  Regular           &71.46 &65.76 &89.53 &75.83  \\
            &                               & AGLA               &\textbf{72.90} &66.89 &90.66 &\textbf{76.99}  \\
                            \hline
\end{tabular}
}
\label{tab:minicpm}
\end{table*}

\section{Results on POPE when LVLMs Scale Up}
\label{app_pope_scale}
Table~\ref{tab:llava13} and~\ref{tab:blip13} show experimental results on the POPE dataset~\cite{li2023evaluating} when LLaVA-1.5~\cite{liu2023visual} and InstructBLIP~\cite{dai2023instructblip} extend to larger 13B variants. Notably, \textit{AGLA} consistently boosts model performance in all subsets and settings of the POPE dataset~\cite{li2023evaluating}, confirming its robustness towards different model scales.

\begin{table*}[tp]
\caption{Results on the full POPE dataset with LLaVA-1.5 13B~\cite{liu2023visual}.}
\vspace{-2mm}
\centering
\scalebox{1.1}
{
\begin{tabular}{cllccc|c}
\hline
\textbf{Dataset}          & \textbf{Setting}                  & \textbf{Decoding} & Accuracy $\uparrow$  & Precision & Recall & F1 Score $\uparrow$   \\ \hline
\multirow{6}{*}{COCO}  & \multirow{2}{*}{\textit{Random}}           & Regular           &83.31 &91.46 &73.48 &81.49                                         \\&                               & AGLA               &\textbf{87.26} &97.94 &76.13 &\textbf{85.67}  \\
                          \cline{2-7} 
                          & \multirow{2}{*}{\textit{Popular}}     & Regular           &82.47 &89.55 &73.53 &80.75  \\
                                    &                               & AGLA               &\textbf{86.46} &96.12 &76.00 &\textbf{84.88}   \\

                        \cline{2-7}
                          & \multirow{2}{*}{\textit{Adversarial}} &  Regular           &80.00 &84.46 &73.53 &78.62  \\
                                   &                               & AGLA               &\textbf{84.56} &91.78 &75.93 &\textbf{83.11}  \\

                            \hline
\multirow{6}{*}{VQA}  & \multirow{2}{*}{\textit{Random}}           & Regular           &83.60 &86.95 &79.07 &82.82                                         \\ &                               & AGLA               &\textbf{89.40} &94.57 &83.60 &\textbf{88.74}  \\

                          \cline{2-7} 
                          & \multirow{2}{*}{\textit{Popular}}     &  Regular           &81.16 &82.53 &79.06 &80.76  \\
                                    &                               & AGLA               &\textbf{86.93} &89.57 &83.60 &\textbf{86.48}   \\ 

                        \cline{2-7}
                          & \multirow{2}{*}{\textit{Adversarial}} &  Regular           &76.43 &74.79 &79.73 &77.18  \\
 &                               & AGLA               &\textbf{80.86} &79.22 &83.67 &\textbf{81.38}  \\
            
                            \hline
\multirow{6}{*}{GQA}  & \multirow{2}{*}{\textit{Random}}           & Regular           &84.50 &87.52 &80.46 &83.84                                         \\
             &                               & AGLA               &\textbf{89.43} &93.84 &84.40 &\textbf{88.87}  \\ 
                          \cline{2-7} 
                          & \multirow{2}{*}{\textit{Popular}}     & Regular           &80.67 &80.78 &80.46 &80.62  \\
               &                               & AGLA               &\textbf{86.80} &88.65 &84.40 &\textbf{86.47}   \\ 
                        \cline{2-7}
                          & \multirow{2}{*}{\textit{Adversarial}} &  Regular           &77.10 &75.29 &80.66 &77.88  \\
            &                               & AGLA               &\textbf{82.53} &81.32 &84.46 &\textbf{82.86}  \\
                            \hline
\end{tabular}
}
\label{tab:llava13}
\end{table*}

\begin{table*}[tp]
\caption{Results on the full POPE dataset with InstructBLIP 13B~\cite{dai2023instructblip}.}
\vspace{-2mm}
\centering
\scalebox{1.1}
{
\begin{tabular}{cllccc|c}
\hline
\textbf{Dataset}          & \textbf{Setting}                  & \textbf{Decoding} & Accuracy $\uparrow$  & Precision & Recall & F1 Score $\uparrow$   \\ \hline
\multirow{6}{*}{COCO}  & \multirow{2}{*}{\textit{Random}}           & Regular           &82.50 &86.35 &77.20 &81.52                                         \\&                               & AGLA               &\textbf{88.67} &95.81 &80.87 &\textbf{87.71}  \\
                          \cline{2-7} 
                          & \multirow{2}{*}{\textit{Popular}}     & Regular           &79.30 &81.64 &75.60 &78.50  \\
                                    &                               & AGLA               &\textbf{85.36} &88.86 &80.86 &\textbf{84.67}   \\

                        \cline{2-7}
                          & \multirow{2}{*}{\textit{Adversarial}} &  Regular           &75.96 &76.44 &75.06 &75.74  \\
                                   &                               & AGLA               &\textbf{82.67} &84.02 &80.66 &\textbf{82.31}  \\

                            \hline
\multirow{6}{*}{VQA}  & \multirow{2}{*}{\textit{Random}}           & Regular           &82.16 &82.05 &82.33 &82.19                                        \\ &                               & AGLA               &\textbf{89.53} &90.06 &88.86 &\textbf{89.46}  \\

                          \cline{2-7} 
                          & \multirow{2}{*}{\textit{Popular}}     &  Regular           &76.96 &73.89 &83.40 &78.35  \\
                                    &                               & AGLA               &\textbf{83.73} &80.59 &88.86 &\textbf{84.52}   \\ 

                        \cline{2-7}
                          & \multirow{2}{*}{\textit{Adversarial}} &  Regular           &72.23 &67.74 &84.86 &75.34  \\
 &                               & AGLA               &\textbf{75.66} &70.11 &89.46 &\textbf{78.61}  \\
            
                            \hline
\multirow{6}{*}{GQA}  & \multirow{2}{*}{\textit{Random}}           & Regular           &81.53 &81.70 &81.26 &81.48                                         \\
             &                               & AGLA               &\textbf{87.60} &88.63 &86.26 &\textbf{87.43}  \\ 
                          \cline{2-7} 
                          & \multirow{2}{*}{\textit{Popular}}     & Regular           &71.86 &68.82 &79.93 &73.96  \\
               &                               & AGLA               &\textbf{79.10} &75.45 &86.26 &\textbf{80.49}   \\ 
                        \cline{2-7}
                          & \multirow{2}{*}{\textit{Adversarial}} &  Regular           &71.60 &67.76 &82.40 &74.36  \\
            &                               & AGLA               &\textbf{74.43} &69.88 &85.86 &\textbf{77.05}  \\
                            \hline
\end{tabular}
}
\vspace{-2mm}
\label{tab:blip13}
\end{table*}

\section{Results on Perception-related MME}
\label{app_mme}
To validate the effectiveness of our model for enhancing general perception capability for LVLMs, we perform experiments on the perception-related tasks of the MME benchmark~\cite{fu2023mme}. As shown in Table~\ref{tab:mme_perception}, our model obtains much better performance than regular baselines and achieves uniformly improvement than previous state-of-the-art models. The improvement may come from the ensemble of local and global image attention to capture both discriminative and generative features, which is important for enhancing the visual perceptual abilities of LVLMs.

\begin{table*}[!t]
\caption{Results on MME perception-related tasks.}
\centering
\resizebox{1\linewidth}{!}{%
\begin{tabular}{@{}llllllllllll|l@{}}
\toprule
Model                         & Decoding & \multicolumn{1}{c}{\textit{Existence}} & \multicolumn{1}{c}{\textit{Count}} & \multicolumn{1}{c}{\textit{Position}} & \multicolumn{1}{c}{\textit{Color}} & \multicolumn{1}{c}{\textit{Posters}} & \multicolumn{1}{c}{\textit{Celebrity}} & \multicolumn{1}{c}{\textit{Scene}} & \multicolumn{1}{c}{\textit{Landmark}} & \multicolumn{1}{c}{\textit{Artwork}} & \multicolumn{1}{c|}{\textit{OCR}} & \multicolumn{1}{c}{\textit{\textbf{\begin{tabular}[c]{@{}c@{}}Perception \\ Total\end{tabular}}}} \\ \midrule
\multirow{5}{*}{LLaVA-1.5~\cite{liu2023visual}}     
& Regular  & 175.67 & 124.67 & 114.00 & 151.00 & 127.82& 113.59 & 148.30 & 129.95 & 102.20 & 92.00 & 1279.19 \\
& DOLA  & 175.00 & 108.33 & 90.00 & 138.33 & 121.43 & 108.82 & 146.50 & 124.12 & 107.50 & 112.50 & 1232.54 \\
& OPERA  & 180.67 & 133.33 & 123.33 & 155.00 & 134.69 & 116.76 & 152.75 & 133.01 & 103.25 & 100.00 & 1332.79 \\
& VCD      & 184.66 & 138.33 & 128.67 & 153.00 & 132.11 & 120.94 & 152.20 & 140.45 & 109.60& 104.00 & 1363.96 \\
& AGLA      & \textbf{195.00} & \textbf{153.89} & \textbf{129.44} & \textbf{161.67} & \textbf{137.07} & \textbf{126.96} & \textbf{157.42} & \textbf{160.13} & \textbf{116.08} & \textbf{135.00} & \textbf{1472.66} \\ \midrule
\multirow{5}{*}{InstructBLIP~\cite{dai2023instructblip}}     
& Regular  & 161.67 & 62.22 & 63.33 & 114.44 & 120.41 & 94.80 & 143.00 & 131.74 & 96.92 & 80.83 & 1069.36 \\
& DOLA  & 175.00 & 50.00 & 46.67 & 113.33 & 130.27 & 96.18 & 148.25 & 131.41 & 99.25 & 72.50 & 1062.86 \\
& OPERA  & 175.00 & 55.00 & 50.00 & 118.15 & 122.86 & 80.00 & 149.25 & 138.79 & 90.75 & 65.00 & 1044.80 \\
& VCD      & 170.00 & 61.67 & 57.22 & 114.44 & 121.09 & \textbf{104.41} & 140.75 & \textbf{140.96} & \textbf{103.08} & \textbf{82.50} & 1096.12 \\
& AGLA      & \textbf{180.00} & \textbf{63.33} & \textbf{65.56} & \textbf{119.44} & \textbf{130.38} & 96.57 & \textbf{150.58} & 135.76 & 97.50 & 70.00 & \textbf{1109.12} \\
\bottomrule
\end{tabular}
}
\label{tab:mme_perception}

\end{table*}

\begin{table}[ht]
\caption{An ablation study on different weighting factors $\alpha$ with LLaVA-1.5 7B~\cite{liu2023visual}.}
\centering
\begin{tabular}{lcccc}
\toprule
\multicolumn{1}{l}{$\alpha$}  & Accuracy $\uparrow$ & Precision & Recall & F1 Score $\uparrow$   \\ 
\midrule
0.5         & 81.80    & 86.30     & 75.60  & 80.60 \\
1.0         & 82.83    & 87.80     & 76.27  & 81.63 \\
1.5         & 83.40    & 88.66     & 76.60  & 82.19 \\
2.0         & 83.80    & 89.12     & 77.00  & 82.62 \\
3.0         & 84.11    & 88.53     & 78.37  & 83.14 \\
\bottomrule
\end{tabular}
\vspace{-3mm}
\label{tab:ablationalpha}
\end{table}

\section{Effect of Hyper-parameters}
\label{app_hyper}
\noindent \textbf{Effect of Weighting Factor.} We investigate the influence of the weighting factor $\alpha$ in Eq.~\ref{fuse} on the POPE-COCO dataset with the adversarial setting in Table~\ref{tab:ablationalpha}. Increasing $\alpha$ is equivalent to amplifying the importance of discriminative local features from augmented images, which is important for mitigating object hallucinations, so our model gets consistent improvement when $\alpha$ increases. On the other hand, increasing $\alpha$ can be viewed as applying a smaller temperature to sampling decoding, which can skew the distribution towards high-probability tokens and avoid selecting low-probability random tokens. Nevertheless, introducing augmented images for decoding with different $\alpha$ consistently outperforms the regular decoding, which can validate the stability of our model in mitigating object hallucinations.

\noindent \textbf{Effect of Adaptive Plausible Constraint Factor.} We investigate the influence of the adaptive plausible constraint factor $\beta$ in Eq.~\ref{constraint} on the POPE-COCO dataset with the adversarial setting in Table~\ref{tab:ablationbeta}. Larger $\beta$ indicates more aggressive truncation, keeping only high-probability tokens. The table illustrates that our model is robust to the change of $\beta$. However, the factor is important for generative tasks to avoid generated tokens being dominated by local image features from the augmented images and compromising global perception ability from the original images.

\section{Qualitative Results for Different Queries}
\begin{figure*}[ht]
\centering
\includegraphics[width=\textwidth]{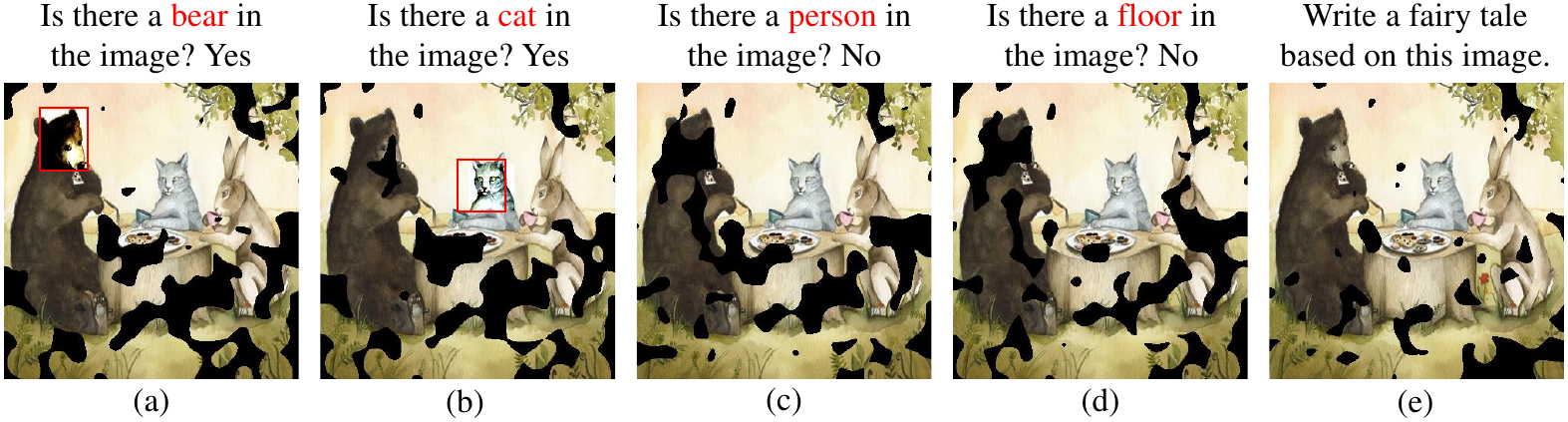}
\vspace{-3mm}
\caption{Masked images with different prompt queries. (a) and (b) are prompts that contain objects in the image, where the detected objects are marked in red. (c) and (d) are prompts that do not contain objects in the image. (e) is a general prompt.} 
\label{match}
\vspace{-3mm}
\end{figure*}

Fig.~\ref{match} shows the masked images generated by our Image-Prompt Matching (IPM) module with different prompt queries. First, for questions that result in ``yes" answers (\ie, Fig. (a) and (b)), the matching model tries to match the objects in the input queries and mask out other irrelevant backgrounds to avoid interference. Second, for questions that result in ``no" answers (\ie, Fig. (c) and (d)), the matching model fails to match any objects and tends to match background areas. So our model can mask some irrelevant objects to mitigate object hallucination caused by object co-occurrence association. Third, for generative tasks with general prompts that contain no objects (\ie, Fig. (e)), our matching model can still identify areas of interest (\eg, animals and the table) due to the pre-training task on caption generation and mask irrelevant backgrounds, which can mitigate interference and prevent loss of details in generated captions. In summary, IPM is effective for different kinds of input queries and helps to mitigate object hallucinations caused by association.

\begin{table}[tp]
\caption{An ablation study on different constraint factors $\beta$ with LLaVA-1.5 7B~\cite{liu2023visual}.}
\centering
\begin{tabular}{lcccc}
\toprule
\multicolumn{1}{l}{$\beta$}  & Accuracy $\uparrow$ & Precision & Recall & F1 Score $\uparrow$   \\ 
\midrule
0.01        & 84.73    & 89.35     & 78.87  & 83.78 \\
0.1         & 84.11    & 88.53     & 78.37  & 83.14 \\
0.2         & 83.90    & 89.39     & 76.93  & 82.69 \\
0.5         & 84.00    & 89.41     & 77.13  & 82.82 \\
1.0         & 83.53    & 89.24     & 76.27  & 82.24 \\
\bottomrule
\end{tabular}
\vspace{-3mm}
\label{tab:ablationbeta}
\end{table}

\section{Qualitative Results for Multiple-object Queries}
\label{app_multiobject}
\begin{figure*}[ht]
\centering
\includegraphics[width=\textwidth]{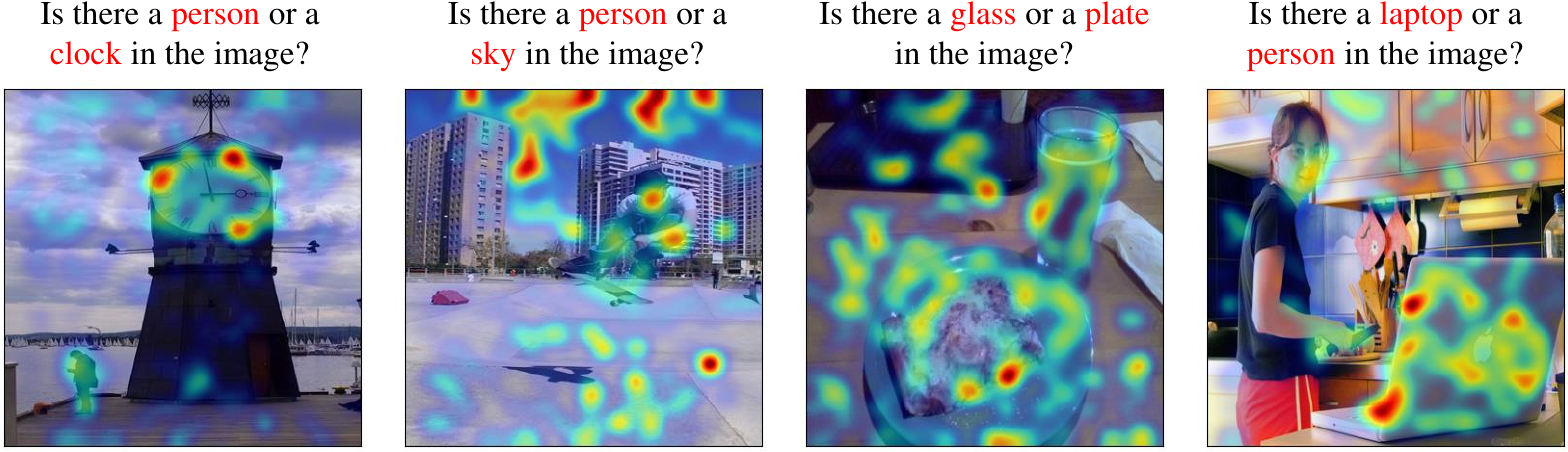}
\vspace{-3mm}
\caption{GradCAM results when queries contain multiple objects.} 
\label{multi}
\vspace{-2mm}
\end{figure*}

In addition to experiments on the ROPE dataset~\cite{chen2024multiobject}, Fig.~\ref{multi} also illustrates that GradCAM can work well when queries contain multiple objects, showing the effectiveness of our model towards multiple-object hallucination mitigation. Furthermore, the experiments also demonstrate that GradCAM can detect general regions of interest rather than a single object, which makes our model competent for a wide range of general tasks in the real world, validated by the improvement over the MME dataset in Table~\ref{tab:mme_perception}.

\section{Prompt for GPT-4 Aided Evaluation}
\label{app_gpt4}
\begin{figure*}[t]
\centering
\includegraphics[width=0.85\textwidth]{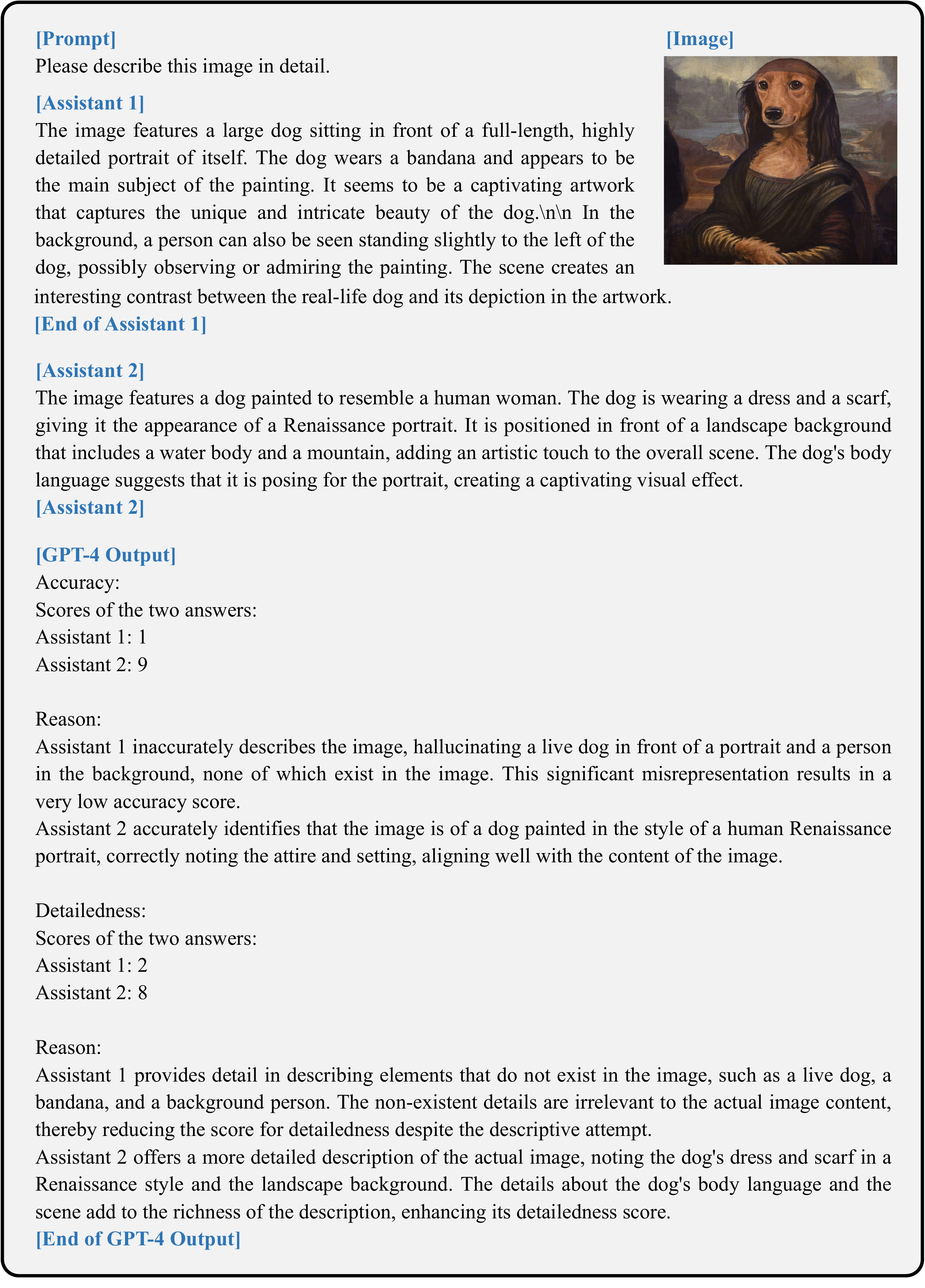}
\caption{An example of the GPT-4 evaluation in open-ended generation tasks. ``Assistant 1'' is Regular decoding and ``Assistant 2'' is the proposed model \textit{AGLA}.}
\label{gpt4example}
\end{figure*}

To evaluate model performance on caption generation, we use GPT-4 to assess the accuracy and detailedness of LVLMs' responses, following previous works~\cite{huang2023opera,vcd}. The prompt used for GPT-4 is detailed in Table~\ref{tab:prompt}. 
Furthermore, we also present an illustrative example in Fig.~\ref{gpt4example} for a better understanding of the evaluation process of GPT-4.

\begin{table}[t]
    \renewcommand\arraystretch{1.0}
    \centering
    \caption{Model performance and inference time on the POPE dataset with LLaVA-1.5~\cite{liu2023visual}.}
    \resizebox{\linewidth}{!}{
    \vspace{-3mm}
    \begin{tabular}{l|cccc}
        \hline
         & VCD & OPERA & AGLA-small & AGLA \\\hline
        \textbf{F1 Score} & 83.16  & 83.55   & 84.11 & 84.58\\\hline
        \textbf{Infer. Time} & 0.56s  & 1.64s & 0.63s & 0.69s\\\hline
    \end{tabular}
    }
    \vspace{-4mm}
    \label{eff}
\end{table}

\section{Efficiency}
We benchmark different models with F1 scores and inference time per sample on the POPE dataset with LLaVA-1.5~\cite{liu2023visual} in Table~\ref{eff}. Experimental results show that our model performs the best with slight inference overhead. We also test \textit{AGLA-small}, an \textit{AGLA} variant using a much smaller matching model with 120M parameters, which is more efficient but achieves a competitive F1 score as well.

\section{Examples for Caption Generation}
\label{app_example}
\begin{figure*}[t]
\centering
\includegraphics[width=0.9\textwidth]{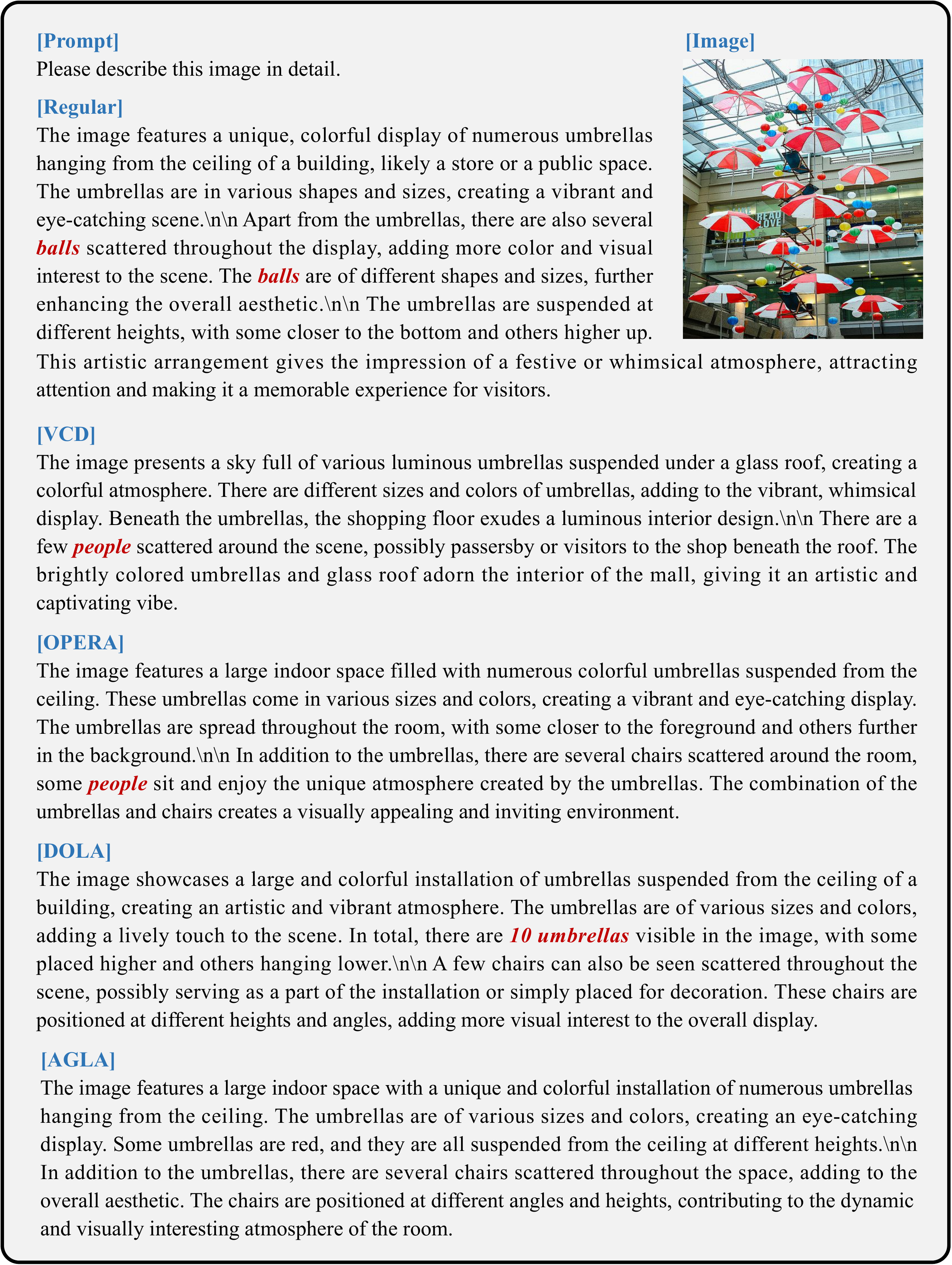}
\caption{An example of generated captions by different decoding methods. Hallucinated content is marked in \textcolor{red}{\textbf{\textit{red}}}.}
\label{chairexample1}
\end{figure*}

\begin{figure*}[t]
\centering
\includegraphics[width=0.9\textwidth]{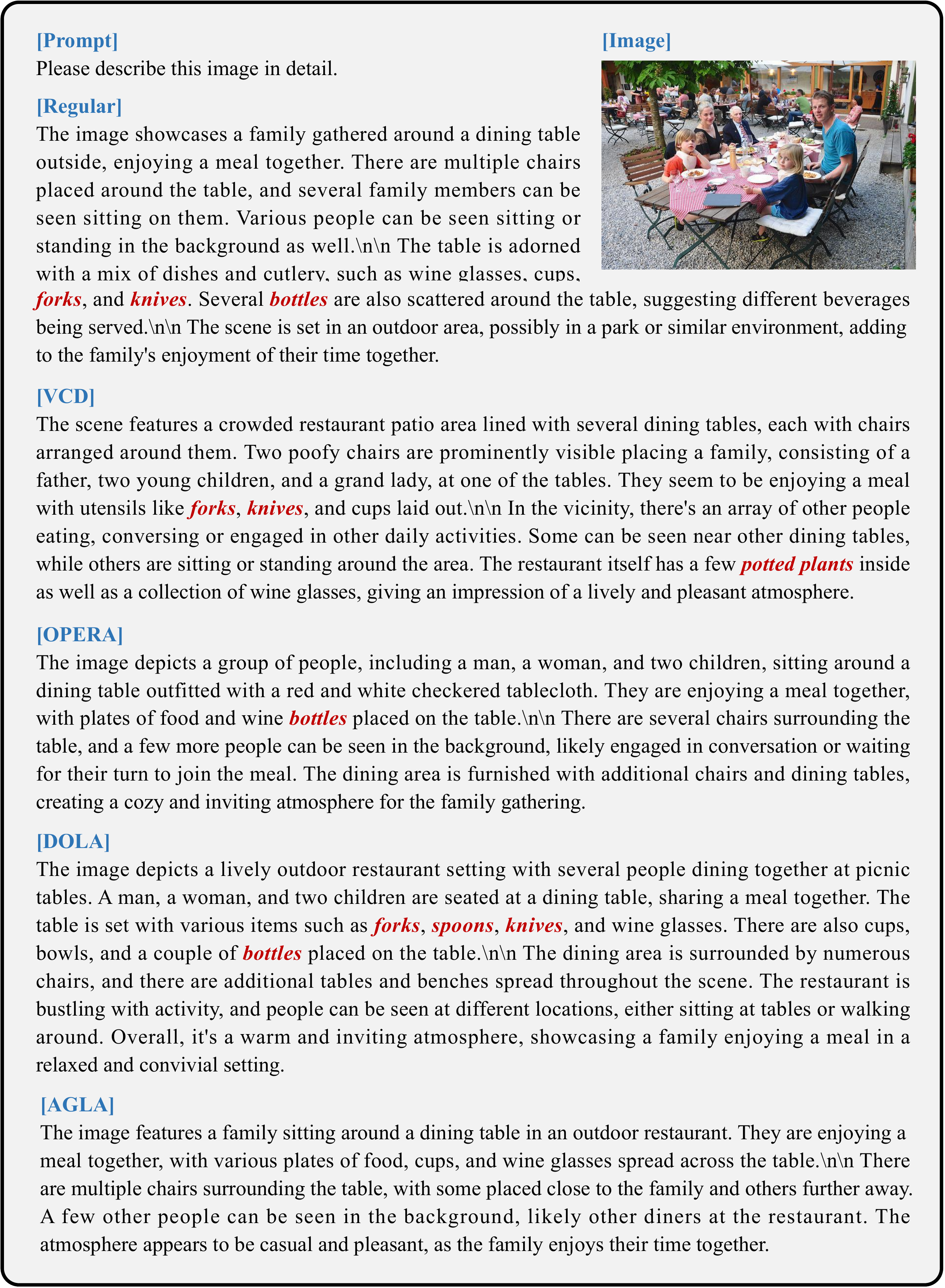}
\caption{An example of generated captions by different decoding methods. Hallucinated content is marked in \textcolor{red}{\textbf{\textit{red}}}.}
\label{chairexample2}
\end{figure*}

In order to demonstrate the quality of generated responses by different models more clearly, we present more qualitative results on the CHAIR evaluation~\cite{chair} in Fig.~\ref{chairexample1} and~\ref{chairexample2}. From the results, we can see that our model can generate captions with fewer object hallucinations, without loss of detailedness of the captions, which is consistent with the results in Table~\ref{tab:chair} and~\ref{tab:gpt4}.

\begin{table*}[ht]\centering
\caption{The prompt for GPT-4 to evaluate captions with Accuracy and Detailedness, following previous works~\cite{huang2023opera,vcd}.}
\begin{minipage}{0.95\textwidth}
\centering
\begin{tcolorbox} 
    \centering
   
      \small
    \begin{tabular}{p{0.95\textwidth}} \hline \\
   \textbf{Description:} \\    
   
   AI that scores image description accuracy and detailedness.

   \\ \midrule
   \\
   \textbf{Instructions:} \\   \\
   
You are an AI designed to evaluate and score the performance of two AI assistants in describing a given image. Your primary focus is on the accuracy and detailedness of their descriptions. You will assess the accuracy by checking for hallucinations - any part of the description that is inconsistent with the image content. For detailedness, you will consider how rich the response is in necessary details, excluding any hallucinated parts. You will provide scores on a scale from 1 to 10 for each assistant separately, based on these criteria. After scoring, you will offer an explanation for your evaluation, ensuring it is free from bias and not influenced by the order of presentation of the responses.
\\ \\
\textbf{Input format:} \\ \\
\lbrack{}Assistant 1\rbrack{}\\
 \{Response 1\}  \\
\lbrack{}End of Assistant 1\rbrack{} \\
\\
\lbrack{}Assistant 2\rbrack{} \\
 \{Response 2\}\\
\lbrack{}End of Assistant 2\rbrack{} \\
\\
\textbf{Output format:}\\
\\
Accuracy:\\
Scores of the two answers:\\
Reason:\\
\\
Detailedness:\\
Scores of the two answers:\\
Reason:\\ \\

\bottomrule
    \end{tabular}
\end{tcolorbox}
    \label{tab:prompt}
\end{minipage}
\end{table*}

\end{document}